\documentclass{article}
\usepackage{jfrExamplee}
\usepackage{graphicx}
\usepackage{apalike}
\usepackage{setspace}
\usepackage{bm}
\usepackage{bbm}
\usepackage{comment}
\usepackage{xcolor}
\usepackage{amsmath}
\usepackage{amssymb}
\usepackage{caption}
\usepackage{pifont}
\usepackage{float}
\usepackage{subfigure}
\usepackage{algorithmicx}
\usepackage[noend]{algpseudocode}
\usepackage{booktabs}
\newcommand*\samethanks[1][\value{footnote}]{\footnotemark[#1]}
\newcommand{\cmark}{ {\color{green} \ding{51}} }%
\newcommand{\xmark}{ {\color{red} \ding{55}} }%
\usepackage{booktabs}
\usepackage{tablefootnote}
\usepackage{tabularx}
\usepackage{makecell}

\DeclareMathOperator*{\argmin}{arg\,min}

\title{The ReSWARM Microgravity Flight Experiments: Planning, Control, and Model Estimation for On-Orbit Close Proximity Operations}

\author{
Bryce Doerr\thanks{Equal contribution.}\\
Department of Aeronautics and Astronautics\\
Massachusetts Institute of Technology\\
77 Massachusetts Avenue, Cambridge, MA 02139\\
\texttt{bdoerr@mit.edu} \\
\AND
Keenan Albee\samethanks\\
Department of Aeronautics and Astronautics\\
Massachusetts Institute of Technology\\
77 Massachusetts Avenue, Cambridge, MA 02139\\
\texttt{albee@mit.edu} \\
\And
Monica Ekal\samethanks\\
Institute for Systems and Robotics\\
Instituto Superior T\'ecnico\\
Avenida Rovisco Pais 1, Lisbon, Portugal \\
\texttt{mekal@isr.tecnico.ulisboa.pt} \\
\And
Rodrigo Ventura\\
Institute for Systems and Robotics\\
Instituto Superior T\'ecnico\\
Avenida Rovisco Pais 1, Lisbon, Portugal \\
\texttt{rodrigo.ventura@isr.tecnico.ulisboa.pt} \\
\And
Richard Linares \\
Department of Aeronautics and Astronautics\\
Massachusetts Institute of Technology\\
77 Massachusetts Avenue, Cambridge, MA 02139\\
\texttt{linaresr@mit.edu} \\
}

\begin{document}

\maketitle

\begin{abstract}
On-orbit close proximity operations involve robotic spacecraft maneuvering and making decisions for a growing number of mission scenarios demanding autonomy, including on-orbit assembly, repair, and astronaut assistance. Of these scenarios, on-orbit assembly is an enabling technology that will allow large space structures to be built in situ, using smaller building block modules. However, like many of these scenarios, robotic on-orbit assembly involves several technical hurdles such as changing system models. For instance, grappled modules moved by a free-flying ``assembler" robot can cause significant shifts in system inertial properties, which have cascading impacts on motion planning and control portions of the autonomy stack. Further, on-orbit assembly and other scenarios require collision-avoiding motion planning, particularly when operating in a ``construction site" scenario of multiple assembler robots and structures. 
Multiple key technologies that address these complicating factors for autonomous microgravity close proximity operations are detailed in this work, in particular: (1) global long-horizon planning, accomplished using offline and online sampling-based planner options that consider the system dynamics; (2) the recently proposed RATTLE information-aware planning framework for on-orbit reconfiguration model learning; and (3) robust control tools to provide low-level control robustness using current system knowledge. These approaches were demonstrated for an autonomous on-orbit assembly use case by the RElative Satellite sWarming and Robotic Maneuvering (ReSWARM) experiments using NASA's Astrobee robots on the International Space Station. Results of the ReSWARM experiments are provided along with significant operational and implementation detail discussing the practicalities of hardware implementation and unique aspects of working with the Astrobee free-flyer robots in microgravity. ReSWARM provides a base set of planning and control tools for robotic close proximity operations, demonstrates them in microgravity, and outlines some of the important hardware aspects that future autonomous free-flyers will need to consider.
\end{abstract}

\section{Introduction}
\label{sec:introduction}

On-orbit robotic assembly and servicing of large, complex space structures is an emerging area of autonomy that can improve space-borne observatories, communications relays, and human space exploration among other areas.  On-orbit servicing is a superset of areas including repairing, refueling, and re-provisioning of spacecraft \cite{saleh2003flexibility}. For instance, there are currently many Earth observing satellites investigating the Earth’s oceans and land masses (e.g. Landsat-7 \cite{goward2001landsat}), whose longevity could be increased by on-orbit servicing and assembly. Specifically, new servicing and assembly technologies are being developed through NASA's On-orbit Servicing, Assembly, and Manufacturing 1 (OSAM-1) servicing mission which will refuel Landsat-7 \cite{reed2016restore,coll2020satellite}.

Concepts in the area of robotic assembly of space structures have also been proposed by industry and academia. Tethers Unlimited, Inc. is enabling the area of on-orbit fabrication including antennas, solar panels, and truss structures using SpiderFab \cite{hoyt2013spiderfab}. Made In Space, Inc. is developing the manufacturing and assembly of spacecraft components on-orbit using its Archinaut One concept \cite{patane2017archinaut}. These assemblers are defined by capturing the structure with a robotic arm and climbing the structure for servicing or construction. Alternatively, space structures can be assembled using proximity operations with free-flyer robots, providing benefits in terms of mission flexibility and range of motion \cite{jewison2014definition}. MIT's Space Systems Laboratory (SSL) has used the Astrobee \cite{bualat2015astrobee} and SPHERES robots --- six degree of freedom (DOF) robotic free-flyers operating aboard the International Space Station (ISS) --- to develop capabilities relating to autonomy, multi-agent coordination, and microgravity manipulation \cite{doerr2021safe,ekal2021online,Miller2000}. Since being deployed to the ISS in 2019, Astrobee has begun operations to replace the SPHERES satellites, providing the functionality necessary to demonstrate capabilities in modular motion planning strategies for on-orbit assembly. By enabling the autonomous robotic assembly of space structures and more, on-orbit assembly serves as a useful scenario for other on-orbit autonomy domains including servicing. The sum total of these emerging areas of microgravity robotic autonomy will lead to improved science-gathering (e.g. upgrading new components), mission life extension (e.g. repairing, refueling, and re-provisioning), and other benefits. As such, with the growing availability of on-orbit general-purpose computing power and robotic platforms, the desire to apply autonomy to these scenarios becomes appealing from an efficiency and safety standpoint.

This work addresses the motion planning and control for on-orbit robotic assembly in the presence of inertial parametric uncertainty caused by payload manipulation using an online, adjustable information-aware planning algorithm. Although inertial properties of the payload (e.g. 3D printed parts) may be known, the manipulated system (i.e. the 3D printed part and robotic assembler system) may have inertial parametric uncertainty due to uncertainties in which the payload is grasped. Since an online, adjustable information-aware planning algorithm is a necessary component of many on-orbit robotic autonomy mission scenarios, the method is advances upon a previously proposed motion planning and control framework which used offline-computed informative paths to mitigate uncertainty caused by payload manipulation \cite{doerr2021safe}. Building on this prior work, microgravity experimental results using the Astrobee robotic free-flyer are presented in this context for the first time. Test cases that are representative of on-orbit proximity operations are considered, in particular, a payload transportation or on-orbit assembly scenario where the robot has to transport a payload with unknown inertial properties to the assembly location. The test is composed of the following stages: first, a spacecraft (Astrobee) collects a payload. This results in an increase in the inertial uncertainty of the spacecraft system. Next, the spacecraft mitigates the inertial uncertainty using online information-aware planning and estimation. Finally, the spacecraft precisely deposits the payload in its designated location. A number of planning and control methods, repurposable to other on-orbit autonomy tasks, are discussed in terms of these test cases; techniques include high-level motion planning with obstacle avoidance, robust tube model predictive control (MPC), and online inertial model learning. Implementation and testing on the Astrobee hardware provides lessons learned for the deployment of these motion planning and control techniques in a microgravity robotic system with processing capabilities similar to that of emerging space robotic systems. Algorithmically, a safe and uncertainty-aware design for the assembling of space structures on-orbit is provided which can be expanded to constructing next generation telescopes, space stations, and satellites.

\textit{Statement of Contributions}:
This work provides a number of contributions, focusing on practical implementation and field testing of motion planning and control to on-orbit autonomy relevant for robotic free-flyers:
\begin{itemize}
    \item Providing motion planning and control components for microgravity robotic free-flyer autonomy, with novelties beyond \cite{doerr2021safe} including online information-aware planning and real-time global planners.
    \item Demonstration of the above for an on-orbit payload transportation use case scenario in microgravity experiments, including individual demonstrations of each major planning/control component over a series of two ISS experiments using NASA's Astrobee free-flyer.
    \item Discussion of extensive hardware implementation details and challenges emerging from field testing, with recommendations for general implementation on other systems. 
\end{itemize}

The paper is organized as follows. Section \ref{sec:related_work} discusses relevant prior work for microgravity robotic autonomy; Section \ref{sec:formulation} describes the formulation of the essential planning under uncertainty problem with practical implications for running on hardware; Section \ref{sec:methods} explores key methods and techniques in addressing the problem formulation; Subsection \ref{sec:approach} reveals two planning and control architectures for addressing the problem; Section \ref{sec:results} details the on-orbit experimental results of these architectures during the ReSWARM-1 and ReSWARM-2 ISS test sessions; finally, Section \ref{sec:conclusion} provides discussion and concluding remarks of the hardware-demonstrated methods and future work. 

\section{Related Work}
\label{sec:related_work}
The on-orbit assembly problem touches on multiple areas of active areas of work in the robotics and space systems literature, including real-time motion planning, planning and control under uncertainty, and online system identification. Since on-orbit assembly involves autonomous manipulation of payloads, surveying, robotic locomotion, and safety planning/control it is inherently a robotics problem with unique considerations because of the safety-critical nature of its operating environment. This section addresses these broad areas of research in the robotics literature, highlighting topics which are most relevant to the on-orbit assembly problem.

\subsection{On-Orbit Assembly}
The autonomous on-orbit assembly problem concerns multiple topics, including mission concepts \cite{gralla2007strategies,izzo2005mission}, path planning \cite{badawy2008orbit} as well as control of assembly vehicles and components \cite{mcinnes1995distributed,mcquade1997autonomous}. Works such as \cite{shi2016robust} have addressed the issue of change in inertial parameters during assembly, and designed robust control approaches to deal with this. In contrast, this work considers a free-flyer satellite tasked with carrying components to their assembly location. Optimal and dynamically feasible motion plans are determined and tracked with robust control, while the change in inertial properties from grappling payload is addressed through information-aware motion planning and subsequent parameter estimation.

Aside from progress in on-orbit assembly, significant developments have been made in assembly of structures in terrestrial environments. Specifically, modular control strategies utilizing component geometry have been developed to build structures using ground and aerial robotics \cite{petersen2011termes,willmann2012aerial}. Collaborative multi-robotic systems have also been designed to transport and manipulate voxels to construct plates, enclosures, and cellular beams \cite{jenett2019material} and through coarse and fine manipulation techniques \cite{dogar2015multi}. Distributed control has been used for larger multi-robotic systems so agents can collectively climb and assemble the structure they are building \cite{werfel2014designing}. Likewise, a number of architectures have been proposed for on-orbit assembly strategies. Superquadric potential fields have been applied to assemble components of a structure while providing collision avoidance \cite{badawy2008orbit}. Other methods focus on the specifics of the planning and control stacks required, including innovative techniques from terrestrial robotics such as sampling-based motion planners, nonlinear model predictive control (NMPC), and many others \cite{perez2012lqr,geraerts2007creating,sathya2018embedded,doerr2020motion}.

\subsection{Robotic Motion and Path Planning}
Ground (terrestrial) robotics has provided technological advancements in motion planning and trajectory optimization, which is the foundation necessary to autonomously assemble structures. Gradient-based optimization tehcniques like the Covariant Hamiltonian Optimization for Motion Planning (CHOMP) \cite{ratliff2009chomp} have been popularized for optimal trajectory planning under constraints, constructing near-optimal trajectories based on the cost, dynamics, and obstacle constraints from an initial and possibly unfeasible trajectory. However, even with gradient information such techniques may get stuck in local minimums or suffer under sufficiently complex/cluttered environments and dynamics. Numerically robust and computationally efficient methods to trajectory optimization have been achieved through the use of sequential convex programming (SCP), such as TrajOpt and GuSTO \cite{schulman2013finding} \cite{Bonalli}. Such methods uses sequential convex programming (SCP) to numerically optimize $L_1$ distance penalties for both inequality and equality constraints, solving an approximate convex problem using sequential quadratic programming, sometimes incorporating guarantees from optimal control. A computationally simpler algorithm to SCP is the proximal averaged  Newton-type  method  for optimal control (PANOC) algorithm \cite{stella2017simple}. The PANOC optimizer is a line-search method that integrates Newton-type steps and forward-backward iterations over a real-valued continuous merit function, achieving fast convergence using first-order information of the cost function and a reduction in matrix operations when compared to SCP \cite{stella2017simple,sathya2018embedded}.

Another possibles set of approaches are gradient-free,  like Stochastic Trajectory Optimization for Motion Planning (STOMP) \cite{kalakrishnan2011stomp}. Since STOMP uses no gradient information, costs that are non-differentiable and non-smooth can still be used to find optimal trajectories; however, optimizing trajectories with this technique is inefficient compared to CHOMP and gradient-based approaches because of the loss of cost function information.

Other major planning approaches include the sampling-based planners (SBP). Among them, the rapidly exploring random tree (RRT) and RRT* may be adapted to plan dynamically feasible trajectories \cite{karaman2011sampling}. These algorithms produce probabistically complete solutions, and their optimizing forms contain asymptotically optimal motion planning solutions. Historically, the cost-to-go heuristic for optimizing variants has been a Euclidean distance metric to minimize the $L_2$ distance, but this has been improved to reflect the dynamics and control of systems using cost-to-go pseudo-metric like that of the linear quadratic regulator (LQR) \cite{perez2012lqr}. The advantage of RRT*-based algorithms is that they can be applied to real-time systems in a computationally efficient manner in which the sampling itself can be interrupted at set time-intervals to allow for real-time operation as well as providing explicit collision-checking for collision-free trajectories.

\subsection{Motion Planning under Uncertainty and Model Improvement}
Uncertainty can be caused by effects intrinsic to the system (e.g., unmodeled dynamics, sensor noise, etc.), or external sources (e.g., disturbances like wind, solar pressure, etc.). As robotic systems and the tasks that they are entrusted with get increasingly complex, dealing with this uncertainties has become key to many motion planning schemes. Some of these uncertainties are parametric and can be learned through explicit action of the robotic system; others are unstructured, providing disturbance terms that must be countered through control or replanning. Model-based control, path planning and control with optimal fuel or energy consumption, behaviour prediction, and fault detection are just some situations that require are touched upon by introduction of uncertainty into the model; parametric uncertainty reduction (e.g., moments of inertia), also serves to benefit many of these areas. This is especially relevant for on-orbit assembly operations, where robotic free-flyers would need to precisely and safely transport various uncertain components to their assembly locations while dealing with unstructured uncertainties.

Approaches for inertial uncertainty reduction can be classified into three broad groups based on whether the learned inertial properties are used in planning and control. The first approach is the use of system identification, where optimal excitation design is used to adequately characterise the system. The estimated parameters are then used by the controller for precise task execution. The Newton-Euler equations of motion, or the equations of angular momentum conservation are often a starting point for formulating the estimation problem \cite{keim2006spacecraft,ma2008orbit,rackl2013parameter,wilsona2004mcrls,yoshida2002inertia}.

The second approach can be termed as disturbance rejection, where the effects of uncertainty are treated as bounded, unwanted disturbances, and control algorithms are designed to execute tasks despite them Robust model predictive control, the use of funnel libraries, direct adaptive control and model-free reinforcement learning can be classified as disturbance-rejecting \cite{mayne2005robust,lopez2019dynamic,majumdarFunnelLibrariesRealtime2017,wu2020reinforcement}.

The third group of methods can be addressed as simultaneous learning, where the parameters are learned online and the updated parameters are used in planning and control. In case learning the system parameters is relevant for the task at hand, simultaneous learning approaches can potentially reduce the time and energy consumed by a full system identification, while at the same time also completing the robot's primary task without being too conservative. Indirect adaptive control, learning-based model predictive control, model-based reinforcement learning and active learning techniques fall under this category \cite{espinoza2017concurrent,ostafew2016learning,slade2017simultaneous,webb2014online,albee2022rattle}. 

The principle of active learning is to choose data from which to learn, in order to obtain a greater degree of accuracy and efficiency in the learning process \cite{settles.tr09}. For instance, information-seeking motion plans are often used in conjunction with parameter or state estimation algorithms with the goal of gathering information-rich measurements to improve estimation accuracy. Depending on how the objective is formulated, active-learning approaches could be implicit or explicit. Most implicit class of methods solve an approximation of the stochastic optimal
control problem. The resulting control policy is thus inherently information-seeking. Examples are work on Partially Observable Markov Decision Processes (POMDPs) with covariance minimization \cite{webb2014online} and covariance steering \cite{Okamoto2019}. The scalability of these approaches and their implementation on hardware remains a challenge. Explicit active learning approaches incentivize information-seeking behaviour, for instance using metrics such as Fisher information \cite{wilson2015maximizing}.

Machine learning tools such as Gaussian process regression have been used in literature for disturbance modeling \cite{ostafew2016learning} or for learning non-linear system models \cite{lee2017gp}. For learning inertial parameters, batch or recursive least square methods can be used, provided an estimation problem linear in terms of the estimates can be formulated or approximated, and that information rich exciting trajectories are used to ensure that the regressor is full-rank and invertible \cite{keim2006spacecraft,yoshida2002inertia}, or non-linear least squares optimization can be used. In these cases, \cite{wensing2017linear} provides a formulation for enforcing physical feasibility constraints on the estimates. Sequential filtering methods such as the Kalman Filter or Unscented Kalman Filter can be used for either parameter estimation, or joint estimation of states and parameters \cite{lichter2004state,vandyke2004unscented}. 

\subsection{Model Predictive Control}
Model predictive control (MPC) is a control technique which is particularly notable for its ability to approximately optimally control general dynamical systems under constraints. Model predictive control's used is widespread in robotics due to its modeling flexibility and, recently, its ability to be used in real-time and sometimes with mathematical guarantees on performance. While proofs for MPC stability under certain conditions \cite{mayneRobustModelPredictive2011} exist, guarantees for stochastic systems are generally lacking. Tube MPC is a notable exception, providing robustness guarantees when bounded additive uncertainty is encountered in the system dynamics \cite{rakovicHandbookModelPredictive2019}. Safety guarantees for control are also emerging in other forms, such control barrier functions and reachable set analysis \cite{lopezAdaptiveSafetyUncertain2020} \cite{majumdarFunnelLibrariesRealtime2017} \cite{brunkeSafeLearningRobotics2021a} \cite{singhRobustOnlineMotion2017}.

Tube MPC is a flexible robust control approach for handling unstructured uncertainty. With Tube MPC, a portion of control authority is reserved for robust actuation, often in a simple feedback form to counter disturbances. The guarantee obtained is one of tube robustness---if a system starts in a tube of possible states, it remains within a tube around a nominal trajectory. These tubes are formed around the nominally planned MPC trajectory, and the motion of the system can be thought of as a composition of these planned ``safe sets". Tube MPC methods exist for both nonlinear \cite{Lopez2019} and linear \cite{limonDesignRobustTubebased2008} systems with additive bounded uncertainty---for the on-orbit assembly problem linear tube MPC is of interest for the linear translational satellite (double integrator) dynamics. In essence, two controllers must be produced: (1) a nominal MPC, operating under a modified set of constraints to account for worst-case uncertainty; (2) an ancillary or ``disturbance rejection" controller that provides robustness to aleatoric uncertainty. Given a reference trajectory, tube MPC will repeatedly plan the nominal trajectory online, and append ancillary controller actuation to the system inputs, resulting in tube robustness.

\section{Problem Formulation}
\label{sec:formulation}
The on-orbit assembly problem is now framed for a microgravity free-flyer robotic system. The discussion here highlights this particular robotic system since it has broad applicability to on-orbit assembly problems, serving as a highly mobile robotic base that can grapple and move payloads, observe structures, and more. The key uncertainty sources are introduced---primarily inertial uncertainty for close proximity operations/on-orbit assembly scenarios---and the distinctions between two main types of uncertainty are made. Finally, a sketch of the on-orbit assembly task used for on-orbit experimentation is outlined.

\subsection{System Dynamics and Introducing the Motion Planning Problem}
A free-flyer must move from $\mathbf{x}_0$ to $\mathbf{x}_f$,with continuous-time nonlinear dynamics given by

\begin{equation}
    \dot{\mathbf{x}}(t)=f(\mathbf{x}(t),\mathbf{u}(t),\boldsymbol{\theta}(t)),
    \label{eq:bryce_dyn}
\end{equation}
where $\mathbf{x}(t)\in\mathbb{R}^{n_{\mathbf{x}}}$ is the robot state, $\mathbf{u}(t)\in\mathbb{R}^{n_{\mathbf{u}}}$ is the robot control input, and $\boldsymbol{\theta}(t)\in\mathbb{R}^{n_{\boldsymbol{\theta}}}$ is an uncertain parameter vector about time $t$ and where $n_{\mathbf{x}}$, $n_{\mathbf{u}}$, and $n_{\boldsymbol{\theta}}$ are the sizes of the state, control, and inertia parameter vectors, respectively. The free-flyer robotic system is described about a full 6 degree of freedom (DOF) in translation and rotational motion. The system can be controlled about the 6 DOF with the necessary forces and torques applied to the robot. The inertial parameters contain both the mass and moments of inertia of the system.

\begin{figure}[hbtp!]
  \centering
  \includegraphics[width=0.36\linewidth]{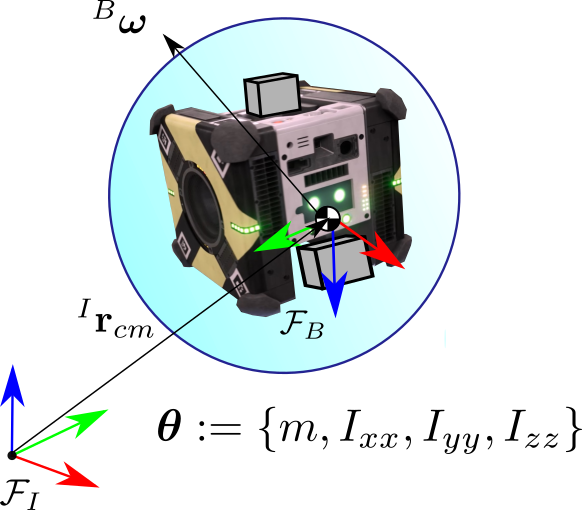}
  \captionsetup{labelfont={bf}}
  \caption{The Newton-Euler dynamics with parameters of interest for the cargo maneuvering scenario.} 
  \label{fig:dyn_cargo}
\end{figure}

The dynamics are depicted in Fig. \ref{fig:dyn_cargo}. A large portion of uncertainty in this scenario is epistemic (see Section \ref{sec:unc})---there is a good system model and the parameters are learnable. (Note that the free-flying dynamics \cite{dubowskyKinematicsDynamicsControl1993} are a possible addition, though for now this work considers only the rigid body dynamics without a manipulator arm.)

The Newton-Euler dynamics for a 6 DOF rigid body expressed in a frame not coincident with the centre of mass (CM), for instance, in a body-fixed frame with origin $B$, can be expressed as
\begin{align}
  \begin{split}
    \begin{bmatrix}
    \mathbf{F} \\ \bm{\tau}
    \end{bmatrix}&= 
    \begin{bmatrix}
    m\mathbb{I}_3 & -m [\mathbf{c}]_{\times}\\  m [\mathbf{c}]_{\times} & \mathbf{I}_{B} 
    \end{bmatrix}
    \begin{bmatrix}
    \dot{\mathbf{v}} \\ \dot{\bm{\omega}}
    \end{bmatrix} +
    \begin{bmatrix}
    ~~m [\bm{\omega}]_{\times}[\bm{\omega}]_{\times} \mathbf{c}~~ \\
    ~~[\bm{\omega}]_{\times} \mathbf{I}_B  \bm{\omega}~~  
    \end{bmatrix},
  \end{split}
  \label{eqn:dyns}
\end{align}

where ${\mathbf{v}}$, $\bm{\omega} \in \mathbb{R}^3$ denote the linear velocity and angular velocity of $B$, $m$ is the system mass, and $\mathbf{c} \in \mathbb{R}^3$ is the CM offset from $B$. The inertia tensor about $B$ is denoted by $\mathbf{I}_{B}$, and it relates to inertia about the CM, $I_{CM}$ as 
\begin{equation}\mathbf{I}_B = \mathbf{I}_{CM} - m[\mathbf{c}]_{\times}[\mathbf{c}]_{\times}\end{equation} 
 $\mathbf{F}, \bm{\tau} \in \mathbb{R}^3$ are the forces and torques applied through the $\mathcal{F}_{B}$ body frame, where $\mathcal{F}_{I}$ indicates the inertial reference frame \cite{ekal2021online}. The identity matrix is denoted by $\mathbb{I}$, while $[-]_{\times}$ indicates a cross product or skew-symmetric matrix.

The inertia tensor $\mathbf{I}$ is given fully as

\begin{align}
  \mathbf{I} &= \begin{bmatrix}
  I_{xx} & I_{xy} & I_{xz}\\
  I_{yx} & I_{yy} & I_{yz}\\
  I_{zx} & I_{zy} & I_{zz}
  \end{bmatrix}.
\end{align}

The inertia tensor may often be assumed to be diagonal if the principal axes are aligned with the base frame, resulting in the diagonal principal moments of inertia $\{ I_{xx}, I_{yy}, I_{zz} \}$.\footnote{Further, it may be shown that only the ratios of the principal moments of inertia (such that $\mathbf{I}$ is diagonal) need be known for Euler's equations: $p_x = \frac{I_{yy} - I_{zz}}{I_{xx}}$, $p_y = \frac{I_{zz} - I_{xx}}{I_{yy}}$, and $p_z = \frac{I_{xx} - I_{yy}}{I_{zz}}$.} 

To develop the motion planning and control for hardware testing, discrete dynamics are defined and also require linear time-varying (LTV) system approximations. First, a first-order Taylor series expansion about an operating point for linearization is determined \cite{markley2014fundamentals}, and then, discretization using a fourth-order Runge-Kutta method with a zero-order hold on control is applied \cite{van1978computing}. This results in an LTV system given by

\begin{equation}\label{axbu}
    \delta\mathbf{x}_{k+1}=A_k\delta\mathbf{x}_k+B_k\delta\mathbf{u}_k,
\end{equation}
where  $A_k$ and $B_k$ are the state and control matrices at a time-step $k$. The terms $\delta\mathbf{x}_k=\left( \mathbf{x}_k-\bar{\mathbf{x}}_k\right)$ and $\delta\mathbf{u}_k=\left( \mathbf{u}_k-\bar{\mathbf{u}}_k\right)$ are the perturbation state and control about some operating point $\bar{\mathbf{x}}_k$. For this work, the operating point occurs at some target $\bar{\mathbf{x}}_k=\mathbf{x}_{des}$.

Ideally, this system will also obey some sort of optimality in its motion, which can be captured in an objective function. The objective of the nominal system (ignoring parametric uncertainty) is to plan and control the robotic free-flyer using a quadratic cost function
\begin{equation}\label{costlqr}
J(\delta{\bf x}_0,\delta U_{0:N-1})=\sum_{k=0}^{N-1}l(\delta{\bf x}_k,\delta{\bf u}_k)+l_N(\delta{\bf x}_N),
\end{equation}
where $\delta U_{0:N-1}=\left[\delta{\bf u}_{0},\delta{\bf u}_{1},\cdots,\delta{\bf u}_{N-1}\right]$ is the control sequence, $l(\delta{\bf x}_k,\delta{\bf u}_k)$ is the running cost, and $l_N(\delta{\bf x}_N)$ is the terminal cost to a time $N$. This is given by
\begin{equation}\label{immediateterminallqr}
l(\delta{\bf x}_k,\delta{\bf u}_k)=\frac{1}{2}\left[ \begin{array}{c} 1 \\\delta { \bf x}_k \\ \delta\mathbf{u}_k \end{array} \right]^{T} \begin{bmatrix} 0& \mathbbm{q}_{k}^{T} & \mathbbm{r}_{k}^{T} \\ \mathbbm{q}_{k}&Q_{k} & P_{k} \\\mathbbm{r}_{k}&P_{k} & R_{k}  \end{bmatrix}\left[ \begin{array}{c} 1\\ \delta{ \bf x}_k \\ \delta\mathbf{u}_k \end{array} \right],\hspace{12pt}l_N(\delta{\bf x}_N)=\frac{1}{2}\delta{ \bf x}_N ^{T}Q_{N}\delta{ \bf x}_N+\delta{ \bf x}_N ^{T}\mathbbm{q}_{N},
\end{equation}
where $\mathbbm{q}_{k}$, $\mathbbm{r}_{k}$, $Q_{k}$, $R_{k}$, and $P_{k}$ are the running weights (coefficients), and $Q_{N}$ and $\mathbbm{q}_{N}$ are the terminal weights. The weight matrices, $Q_{k}$ and $R_{k}$, are positive definite and the block matrix $\begin{bmatrix} Q_{k} & P_{k} \\P_{k} & R_{k}  \end{bmatrix}$ is positive-semidefinite \cite{13}. This is constrained to dynamics and obstacles (structural components) of the assembly problem itself.

\paragraph{Constraints for the Free-Flyer On-Orbit Assembly Scenario}
The constraint sets $\mathcal{X}$ and $\mathcal{U}$ define admissible values for $\mathbf{x}$ and $\mathbf{u}$. For both $\mathbf{x}$ and $\mathbf{u}$, the constraints of interest are

\begin{align}
  &\text{position: }\mathbf{r} \in \mathcal{X}_{module}\\
  &\text{velocity: }\mathbf{v}_{max} < \mathbf{v} < \mathbf{v}_{max}\\
  &\text{angular velocity: }-\bm{\omega}_{max} < \bm{\omega} < \bm{\omega}_{max}\\
  &\text{force input: }\mathbf{f} < \mathbf{f}_{max}\\
  &\text{torque input: }\bm{\tau} < \bm{\tau}_{max}
\end{align}

where the position/obstacle constraint $\mathbf{r} \in \mathcal{X}_{module}$ is confined about the experimental module with obstacles and must be enforced by a form of collision-checker, as mentioned in Section \ref{sec:collisionchecking}. Note that the input constraints are significantly complicated by the coupled nature that inputs are applied through the system's mixer \cite{Jewison2014}. A simplifying assumption is to make each respective constraint small enough that force and torque can independent saturate and not cause individual thruster saturations.

\subsection{Forms of Model Uncertainty and the Stochastic Planning Problem}
\label{sec:unc}

A robotic system has a state vector,  $\mathbf{x} \in \mathbb{R}^{n}$ which describes its configuration. A nominal, deterministic model, $\mathcal{M}$ governs how this state vector evolves,

\begin{align}
  \mathcal{M} : \left\{ \mathbf{x}, \mathbf{u}\right\} \in \mathbb{R}^n, \mathbb{R}^m \rightarrow \mathbf{x} \in \mathbb{R}^n
\end{align}

where inputs $\mathbf{u} \in \mathbb{R}^m$ are available to influence the system's motion.
A mobile robotic system, like a free-flyer transporting payload, will experience uncertainty in its motion model in two broad forms: parametric and unstructured uncertainty. While deterministic models are very useful for proposing nominal solutions to planning and control problems---as in the discussion of the prior section---they do not adequately capture one's knowledge of the model. Uncertainty enters the problem in one of two ways:

\begin{itemize}
  \item \textbf{Epistemic}: Systematic uncertainty, due to information that could be known, but is not in practice. It is a design decision whether or not to incorporate or improve this information.
  \item \textbf{Aleatoric}: Statistical uncertainty, due to unknowns that vary each time an experiment is run.
\end{itemize}

Uncertain parameters are a common form of epistemic uncertainty, represented by a parameter vector $\bm{\theta} \in \mathbb{R}^{n_{\boldsymbol{\theta}}}$. It is necessary to describe knowledge of these parameters in some way---following a Bayesian approach, a belief (a probability distribution) is frequently assigned to the parameters to describe the current level of knowledge,

\begin{align}
  \bm{\theta} \sim \mathcal{N}(\bar{\bm{\theta}}, \bm{\Sigma}_\theta)
\end{align}

where $\bar{\bm{\theta}}$ is the most likely belief value---the center of the Gaussian distribution---and $\bm{\Sigma}_\theta$ is the parameter covariance matrix. A Gaussian has been selected as the parameter belief representation, a common assumption in the literature for a variety of reasons.\footnote{Among these are the Central Limit Theorem implying that Gaussians are often a good choice for poorly understood distributions and convenient propagation properties of Gaussians (i.e., linear transformations of Gaussians are also Gaussians).} Gaussians are used as the assumed belief state throughout this problem formulation, though it is possible to use other distributions if desired.

Uncertain parameters are assumed constant and unknown with a known functional form in the system dynamics model. The dynamics model with parametric uncertainty becomes

\begin{align}
  \label{eq:epi_dyn}
  \dot{\mathbf{x}} &= f(\mathbf{x},\mathbf{u},\bm{\theta})
\end{align}

equivalent to the dynamics of eq. \eqref{eq:bryce_dyn} where parametric uncertainty is considered. There is a ground truth value for these unknowns but the system does not start off with exact knowledge of the parameters' ground truth values, instead relying on an initial belief for the parameter(s), $\mathcal{N}(\hat{\bm{\theta}}_0, \bm{\Sigma}_{\theta,0})$. In the case where epistemic uncertainty is not learned, for all intents and purposes this uncertainty is aleatoric, where $\mathbf{w}_x$ (the additive uncertainty, or noise) is entirely the result of having an inadequate understanding of model parameters. The system under pure noise uncertainty would look similar, except the uncertainty would be a result of unknowable disturbance sources like environmental perturbations, for instance, if the robot gets jostled a bit while traversing the construction site. For the full aleatoric system with constraints, the complete motion planning problem is

\begin{align}
\begin{split}
  \underset{\mathbf{u}(t)}{\text{min}}\ J &= \mathbb{E}\left[ \int_{t_0}^{t_f}l(\mathbf{x}(t), \mathbf{u}(t)) \ dt + l_N(\mathbf{x}(t_f), \mathbf{u}(t_f)) \right]\\
  &l(t) = \mathbf{x}(t)^\top \mathbf{Q} \mathbf{x}(t) + \mathbf{u}(t)^\top \mathbf{R} \mathbf{u}(t)\\
  &l_N(t) = \mathbf{x}(t_f)^\top \mathbf{H} \mathbf{x}(t_f)\\
  \text{such that}: &\\
  &\quad \dot{\mathbf{x}}(t)=f(\mathbf{x}(t),\mathbf{u}(t),\boldsymbol{\theta}) + \mathbf{w}_x\\
  &\quad \mathbf{w}_x \sim \mathcal{N}(0, \bm{\Sigma}_w)\\
  &\quad \mathbb{E}\left[ \mathbf{x}(t_0) \right] \sim \mathcal{N}(\bar{x}_0, \bm{\Sigma}_{x,0})\\
  &\quad u \in \mathcal{U}\\
  &\quad x \in \mathcal{X}_{free}.
\end{split}
\label{eq:traj_opt}
\end{align}

The parameter(s), $\bm{\theta} $, may be updated enroute if new information characterizing the belief state becomes available, while $\mathbf{w}_x$ is a form of aleatoric uncertainty. 
The problem is a stochastic, constrained, trajectory optimization which in practice is challenging to obtain optimal solutions for. However, a common approach in robotics and controls is assuming a known, zero-mean, Gaussian additive disturbance on the system dynamics, and providing robust or probabilistically robust (i.e., chance-constrained) means of planning and control. However, such approaches frequently overlook the $\textit{dual control}$ problem---that is, acknowledging that these uncertainties can sometimes be better understood $\textit{online}$, as the system executes. Even in this case, for a limited set of systems, optimal solutions can be found; for instance, the stochastic HJB equation leads to the well-known linear quadratic Gaussian (LQG) for this system (without constraints).

\paragraph{Model Parameters and Parametric Uncertainty of the Free-Flyer System}
Returning to the on-orbit assembly free-flyer dynamics, the full parameter vector containing all inertial property uncertainty may be written as

\begin{align}
  \bm{\theta} \triangleq \left[ m, I_{xx}, I_{yy}, I_{zz}, I_{xy}, I_{xz}, I_{yz}, \mathbf{c}_x, \mathbf{c}_y, \mathbf{c}_z \right]^\top.
\end{align}

A simplifying assumption of negligible Center of Mass (CM) offset is made for approximating changes in the model during the Astrobee cargo maneuvering scenario. This simplification also aids the parameter estimation and planning tasks by reducing the dimension of $\bm{\theta}$. The model's parameters, assuming negligible center of mass change and known principal axes, are then

\begin{align}
  \bm{\theta} \triangleq \left[m, I_{xx}, I_{yy}, I_{zz} \right]^\top,
\end{align}

where each of which has an associated initial belief, $\theta_i \sim \mathcal{N}(\bar{\theta_i}, \sigma_{\theta,i})$. It is assumed the uncertainty is primarily due to the unknown parameters $\bm{\theta}$, but additive noise, $\mathbf{w}$, is also possible.

\subsection{The On-Orbit Assembly Scenario}
The on-orbit assembly scenario considers the stochastic motion planning problem for a robotic free-flyer of the prior section, where portions of the scenario might benefit from online parametric model updating. For the on-orbit assembly scenario, three types of planning and control maneuvers are of interest. The first is the use of offline planning that can be produced in passive waiting areas. In this case, the system model knowledge is considered adequate and robust tracking is unnecessary. The second is the use of online planning, control, and estimation that is responsive to poor model knowledge. Therefore, real-time planning is desirable as new model information is learned. Lastly, offline planning that can be produced in passive waiting areas with adequate system model knowledge is considered. Further, the presence of uncertainty in this case makes robust tracking desirable. The experimental setup of the on-orbit assembly scenario is covered in greater detail in Section \ref{sec:setup}; the next section introduces the methods and approaches to cover these areas.\footnote{A portion of these methods are also documented in the thesis \cite{albeeOnlineInformationawareMotion2022}.}

\section{Methods and Approach}
\label{sec:methods}

The on-orbit assembly planning and control problem is tackled through a tiered combination of global and local planning and low-level control. Additionally, information-aware and robust planning and control methods are deployed in some portions of the stack to aid in dealing with system uncertainty. First, the methods deployed are introduced, followed by their composition as a suite of on-orbit assembly under uncertainty tools.

\subsection{Dynamics-Aware and Real-Time Global Planning Methods}
\label{sec:plan}
The motion planning for on-orbit assembly using a free-flyer is developed using RRT* based methods including LQR-RRT* and kinodynamic RRT. Additionally, trajectory smoothing using LQR shortcutting is used to find shortcut trajectories based off the initial sample-based trajectory. The motivation of using sampling-based motion planning is that it provides computationally efficient, collision-free trajectories through random sampling of the complex state-space \cite{perez2012lqr}. In literature, LQR-RRT* and trajectory smoothing have been discussed extensively \cite{paden2016survey,geraerts2007creating,kallmann2008planning}. In this work, LQR-RRT* and trajectory smoothing using LQR shortcutting are discussed with application to on-orbit assembly using free-flyers which follows from previous work \cite{perez2012lqr,doerr2020motion}.
\subsubsection{Optimizing Global Planning Module: LQR-RRT*}
For a nonlinear system given in eq. \eqref{eqn:dyns}, the goal is to obtain a trajectory that minimizes the quadratic cost function given in Eqs. \eqref{costlqr} and \eqref{immediateterminallqr} with initial and final states, $\mathbf{x}_0$ and $\mathbf{x}_{des}$, respectively. The quadratic cost is simplified to
\begin{equation}
\begin{gathered}
J(\delta{\bf x}_0,\delta U_{0:N-1})=\sum_{k=0}^{N-1}\delta{ \bf x}_k ^{T}\mathbbm{q}_{k}+\delta{ \bf u}_k^{T}\mathbbm{r}_{k}+\frac{1}{2}\delta{ \bf x}_k ^{T}Q_{k}\delta{ \bf x}_k+\frac{1}{2}\delta{ \bf u}_k ^{T}R_{k}\delta{ \bf u}_k +\delta{ \bf u}_k ^{T}P_{k}\delta{ \bf x}_k
\\+\frac{1}{2}\delta{ \bf x}_N ^{T}Q_{N}\delta{ \bf x}_N+\delta{ \bf x}_N ^{T}\mathbbm{q}_{N}.
\end{gathered}
\end{equation}
The optimal solution with respect to the cost function in terms of the control sequence is given by
\begin{equation}\label{minJ}
\delta U_{0:N-1}^{\star}(\delta{\bf x}_0)=\argmin_{\delta U_{0:N-1}} J(\delta{\bf x}_0,\delta U_{0:N-1}).
\end{equation}
To solve for the control solution in eq. \eqref{minJ}, a value iteration is used which determines the optimal cost-to-go (value) starting from the final time-step and moving backwards in time minimizing the control sequence. This is given by
\begin{subequations}
\begin{equation}\label{costgo}
J(\delta{\bf x}_{k},\delta U_{k:N-1})=\sum_{k}^{N-1}l(\delta{\bf x}_k,\delta{\bf u}_k)+l_N(\delta{\bf x}_N),
\end{equation}
\begin{equation}\label{valuefunction}
V(\delta{\bf x}_{k})=\min_{\delta U_{k:N-1}} J(\delta{\bf x}_{k},\delta U_{k:N-1}),
\end{equation}
\end{subequations}
which is similar to eq. \eqref{costlqr}
and eq. \eqref{minJ}, but the cost starts from time-step $k$ instead. The optimal cost-to-go at time-step $k$ is a quadratic function given by
\begin{equation}\label{quadfun}
V(\delta{\bf x}_{k})=\frac{1}{2}\delta{\bf x}_k^{T}S_k\delta{\bf x}_k+\delta{\bf x}_k^{T}\mathbf{s}_k+ c_k,
\end{equation}
where $S_k$, $\mathbf{s}_k$, and $c_k$ are computed backwards in time from the final conditions $S_N=Q_N$, $\mathbf{s}_N=\mathbbm{q}_N$, and $c_N=c$. Thus, the minimization of the control sequence becomes a minimization over a control input at a time-step which is known as the principle of optimality \cite{31}. To find the optimal value, the Ricatti equations are propagated backwards from the final conditions as given by
\begin{subequations}
\begin{equation}\label{rit1}
\begin{split}
S_k=A_k^{T}S_{k+1}A_k+Q_k-\left(B_k^{T}S_{k+1}A_k+P_k^{T} \right)^{T}\left(B_k^{T}S_{k+1}B_k+R_k \right)^{-1}\left(B_k^{T}S_{k+1}A_k+P_k^{T}\right),
\end{split}
\end{equation}
\begin{equation}\label{rit2}
\begin{gathered}
\mathbf{s}_k=\mathbbm{q}_k+A_k^{T}\mathbf{s}_{k+1}+A_k^{T}S_{k+1}\mathbf{g}_k
\\-\left(B_k^{T}S_{k+1}A_k+P_k^{T}\right)^{T}\left(B_k^{\mathsf{T}}S_{k+1}B_k+R_k\right)^{-1}\left(B_k^{T}S_{k+1}\mathbf{g}_k+B_k^{T}\mathbf{s}_{k+1}+\mathbbm{r}_k\right),
\end{gathered}
\end{equation}
\begin{equation}\label{rit3}
\begin{gathered}
c_k=\mathbf{g}_k^{T}S_{k+1}\mathbf{g}_k+2\mathbf{s}_{k+1}^{T}\mathbf{g}_k+c_{k+1}
\\-\left(B_k^{T}S_{k+1}\mathbf{g}_k+B_k^{T}\mathbf{s}_{k+1}+\mathbbm{r}_k\right)^{T}\left(B_k^{\mathsf{T}}S_{k+1}B_k+R_k\right)^{-1}\left(B_k^{T}S_{k+1}\mathbf{g}_k+B_k^{T}\mathbf{s}_{k+1}+\mathbbm{r}_k\right).
\end{gathered}
\end{equation}
\end{subequations}
Thus, approximately optimal LQR solutions from nonlinear equations of motion about a quadratic cost function can be found which can be used in conjunction with RRT* \cite{karaman2011sampling} to obtain dynamically feasible continuous trajectories with the asymptotic optimality property. LQR-RRT* consists of seven steps of an algorithm including: 
\begin{itemize}
	\item \textit{Random sampling}: The state-space is randomly sampled uniformly to obtain a node (state $\mathbf{x}_{rand}$).
	\item \textit{Nearest node}: With $\mathbf{x}_{rand}$ and a current set of nodes $\mathbb{N}$ of the tree (trajectory), the nearest node in the tree relative to $\mathbf{x}_{rand}$ is obtained using the value function in eq. \eqref{quadfun} given by
	\begin{equation}
	    \mathbf{x}_{nearest}=\argmin_{\mathbf{x}'\in \mathbb{N}}(\mathbf{x}'-\mathbf{x}_{rand})^TS_{rand}(\mathbf{x}'-\mathbf{x}_{rand})+(\mathbf{x}'-\mathbf{x}_{rand})\mathbf{s}_{rand}+c_{rand},
	\end{equation}
	where $S_{rand}$, $\mathbf{s}_{rand}$, and $c_{rand}$ are computed about the time-step which $\mathbf{x}_{rand}$ occurs.
	\item \textit{LQR steer}: An LQR trajectory is found between nodes $\mathbf{x}_{nearest}$ and $\mathbf{x}_{rand}$. Note that the trajectory found can be a path that moves towards $\mathbf{x}_{rand}$ with a final state $\mathbf{x}_{new}$.
	\item \textit{Near nodes}: Using $\mathbf{x}_{new}$ and set $\mathbb{N}$, a subset of nodes $\mathbb{N}_{near}\subseteq \mathbb{N}$ is found within the vicinity of $\mathbf{x}_{new}$ using eq. \eqref{quadfun} given by
	\begin{equation}
	\resizebox{.9 \hsize}{!}{$ 
	     \left\{ \mathbf{x}'\in \mathbb{N}:(\mathbf{x}'-\mathbf{x}_{new})^TS_{new}(\mathbf{x}'-\mathbf{x}_{new})+(\mathbf{x}'-\mathbf{x}_{new})\mathbf{s}_{new}+c_{new}\leq\gamma\left(\frac{\text{log}n}{n}\right)^{1/n_{\mathbf{x}}}\right\}.$}
	\end{equation}
    \item \textit{Choosing a parent}: LQR trajectories for each node in $\mathbb{N}_{near}$ are found with respect to $\mathbf{x}_{new}$. The node with the lowest cost ($\mathbf{x}_{min}$) and the trajectory $\sigma_{min}$ becomes the parent node of $\mathbf{x}_{new}$. 
    \item \textit{Collision checking}: The trajectory $\sigma_{min}$ is checked against any obstacles within the state-space and is further discussed in the Collision Avoidance subsection.
	\item \textit{Rewire near nodes}: If $\sigma_{min}$ and $\mathbf{x}_{new}$ are collision-free, $\mathbf{x}_{new}$ is added to the set of nodes $\mathbb{N}$, and then attempts are made to reconnect $\mathbf{x}_{new}$ with the set $\mathbb{N}_{near}$ with LQR trajectories if the cost is less than its current parent node.
\end{itemize}
This algorithm provides a single pass of LQR-RRT* which generates collision-free trajectories built from sampling. Sampling through LQR-RRT* may provide jerky and unnatural paths if not enough samples are taken, thus, trajectory smoothing through LQR shortcutting (see Section \ref{sec:shortt}) is applied to reduce this effect and provide collision-free trajectories.

\subsubsection{An Alternative Global Planning Module: Kinodynamic RRT}
Global planning with considerations for real-time use might also be desirable, particularly if replanning is required. First proposed by \cite{lavalleRRTprogressProspects2001}, kinodynamic-RRT (kino-RRT) explores the state space (as in the previous section) rather than configuration space (as used in basic RRT) and enjoys the benefits of explicit evaluation of constraints, including dynamics. Essentially, forward evaluation of the dynamics is incorporated into \texttt{steer} and an appropriate distance metric is developed so that \texttt{nearest} works appropriately. This can be fast enough for online deployment and provide an initial reference solution to other solvers that fail from poor initial guesses. Compared to LQR-RRT*, kino-RRT need not use the approximated quadratic cost function as a distance metric; alternative distance metrics can be used for evaluation of the algorithm. Pseudo-code for the algorithm is provided in Fig. \ref{alg:kino}. An adapted version of Guided-Kino-RRT is considered, proposed in \cite{Albee2019} for a low-dimensional satellite system. Section \ref{sec:primitives} discusses the motion primitive approach that enables fast planning.

\subsubsection{Motion Primitives}

\label{sec:primitives}
The Guided-Kino-RRT algorithm is summarized in Fig. \ref{alg:kino}, which uses a few modifications for sufficient speed to be usable in real-time. First, a simple cost-to-go distance metric, weighted Euclidean distance to the desired $\mathbf{x}_{rand}$ is used,

\begin{align}
  J = ||\mathbf{x} - \mathbf{x}_{rand}||_{\mathbf{w}}
\end{align}

where $\mathbf{w}$ represents variable weightings on desired values of the state space. While not a true approximation of the cost-to-go, this metric is simple and allows for adjustment of different values of the state space based on relative priority for the system of interest.

Second, progress toward minimizing this metric during $\texttt{steer}$ is accomplished by selection of representative motion primitives. Motion primitives are actions $\mathbf{u}(t) : t \in [t_0, t_a]$ where $t_a - t_0$ is a specified action length. Motion primitives reduce the branching factor of selecting an ``optimal'' action that makes the closest approach to $\mathbf{x}_{rand}$; from the infinite set of possible inputs to apply, only a set of inputs that are ``reasonable'' are used. There are numerous methods for choosing motion primitives, and they feature in a number of motion planning schemes that consider system dynamics, but a general rule of thumb is that the motion primitives be representative of the typical motions the system might execute. Note that in practice one must use an ordinary differential equation solver to compute $\mathbf{x}(t_a)$ per action---this might require selecting an additional timestep $dt < t_a$ for the integration scheme.

\begin{figure}
  \begin{algorithmic}[hbtp!]
    \Procedure{Guided-Kino-RRT}{$\mathcal{X},\mathcal{U}$}
    \State $V \gets {\mathbf{x}_{init}}; E \gets \emptyset;$
    \While{$\mathbf{x}_{err} > tol$}
    \State $\mathbf{x}_{rand}\gets \texttt{sampleFree}$
    \State $\mathbf{x}_{nearest}\gets \texttt{nearest}(V, x_{rand})$
    \State $\mathbf{x}_{new}\gets \texttt{guidedSteer}(\mathbf{x}_{nearest}, \mathbf{x}_{rand})$
    \If{\texttt{ObstacleFree}($\mathbf{x}_{nearest}, \mathbf{x}_{new}$)}
    \State $V \gets V \cup {\mathbf{x}_{new}}; E \gets E \cup {(\mathbf{x}_{nearest}, \mathbf{x}_{new})};$
    \EndIf
    \EndWhile
    \State return $G = (V,E);$
    \EndProcedure
  \end{algorithmic}
  \caption[Guided-Kino-RRT.]{Guided-Kino-RRT, used by the replanning-enabled global planning module. $\mathbf{x}_{err}$ is the error from a desired goal state.}
  \label{alg:kino}
\end{figure}

\begin{figure}
  \begin{center}
    \includegraphics[width=0.8\linewidth]{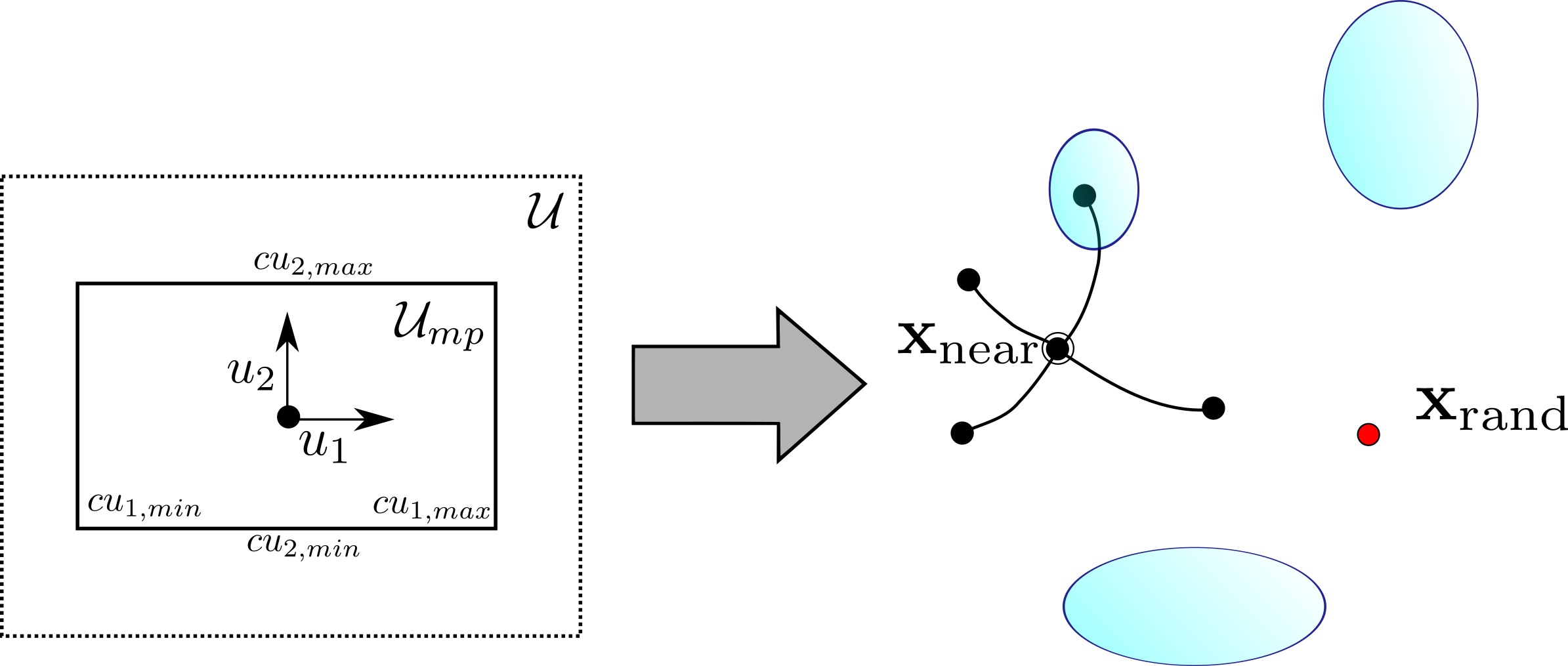}
  \end{center}
  \caption[Motion primitives.]{Motion primitives (right) result from the selection of $\mathcal{U}_{mp}$ (left). Motion primitives must be carefully selected based on state constraints especially considering the density of obstacles (light blue), which will reject unsuccessful expansions.}
  \label{fig:mp}
\end{figure}

Because the goal of the global planner is mainly to enable efficient collision-checking, the translational dynamics are primarily of interest for the free-flyer system. A simple motion primitive selection method yields a surprisingly effective balance between sufficient description of the action set and computational burden. For the input constraints, for each input $u_i \in \mathbf{u}$  select a positive and negative input between the maximum permissible inputs, $u_i^+ = c u_{i, max}$ and $u_i^- = c u_{i, min}$ where $c \in [0, 1]$. The set of motion primitives $\mathcal{U}_{mp}$ is then just the set of all these positive and negative inputs,

\begin{align}
\begin{split}
  &[u_0^-, \dots, 0]^\top,\\
  &[u_0^+, \dots, 0]^\top,\\
  &[0, u_1^-, \dots, 0]^\top,\\
  &[0, u_1^+, \dots, 0]^\top,\\
  &\dots,\\ 
  &[\mathbf{0}]
\end{split}
\end{align}

which will result in an action set of size $2m + 1$, with the addition of the zero ``no action'' option, $[\mathbf{0}]$. $c$ is a scaling option of the max desired motion input which should be chosen conservatively. Additionally, a $t_a$ must be selected for propagation time: small $t_a$ will lead to small $\texttt{steer}$ steps and many more node expansions; large $t_a$ might easily expand nodes beyond typical obstacle density distance, or miss the fidelity of shorter applications of input in making connection decisions, Fig. \ref{fig:mp}. It is emphasized that rough approximations like the above motion primitive selection method above are not an extreme concern because:
\begin{itemize}
  \item Replanning is anticipated---it is not worth creating a time-intensive, perfectly crafted global plan for a nominal set of dynamics.
  \item Trajectory optimization is often performed using either a smoothing approach or a local planner---global plan solutions are effectively guiding initializations with the main goal of enforcing obstacle constraints while remaining dynamically feasible.
\end{itemize}

\subsection{Collision Avoidance}\label{sec:collisionchecking}

In order to prevent collisions between the free-flyer and e.g., an assembled structure, collision avoidance must be considered to remain in $\mathcal{X}_{free}$. Utilizing the advantage of a direct collision checking module is one of the most useful benefits of selecting a sampling-based planner; sampling-based motion planners (e.g. RRT$^*$) include a collision checking feature which is used to generate collision-free trajectories between two nodes (states). If the trajectory between two nodes intersects with an obstacle, the trajectory is thrown out, and a new trajectory is generated at the next iteration. The goal for the on-orbit free-flyer is to plan trajectories for structural assembly under inertial uncertainty while avoiding these obstacles. It is assumed that the obstacles themselves are tracked by either the free-flyer or another spacecraft. By knowing the obstacle states through obstacle tracking, the on-orbit free-flyer can approximate keep-out zones.

 There are a number of possible obstacle representations---one may use ellipsoidal-on-point constraints for instance in order to perform simple collision checking, which are common for e.g., space robotics motion planning scenarios \cite{jewisonModelPredictiveControl2015}. A more sophisticated method unique to sampling-based planners, ellipsoidal-on-ellipsoidal collision checking is also covered here.

\subsubsection{Ellipsoid-on-Point Constraints}
Ellipsoidal obstacle collision checking can be performed via

\begin{align*}
(\mathbf{x} - \mathbf{x}_c)^\top \mathbf{P} (\mathbf{x} - \mathbf{x}_c) \geq 1\\
\mathbf{P} = \begin{bmatrix}
\frac{1}{a^2} & 0 & \\
0 & \frac{1}{b^2} & 0 \\
0 & 0 & \frac{1}{c^2}\\
\end{bmatrix}
\end{align*}
where $\mathbf{P}$ is the ellipsoid shape matrix and $a$, $b$, and $c$ are semi-major and semi-minor axis lengths of the ellipsoid centered at $\mathbf{x_c}$. This is a handy quadratic constraint that can be added on to optimization-based formulations, or used as a black box collision check for sampling-based planning. However, ensuring safety requires expanding the bounding ellipsoid to be sufficiently large to also include the robot's own bounding geometry since the check is only whether a point representation of the robotic system is included within the ellipsoid's volume---this can lead to extremely conservative collision-checking solutions.

This formulation may also be used as a constraint in optimization-based approaches, though it is nonlinear and nonconvex; for sampling-based motion planning, it is a simple check used for generating collision-free segments. Although this work assumes that obstacles are static once they are assembled, dynamic updates to the obstacle state, $\mathbf{x}_{c}$, can also be integrated \cite{albeeOnlineInformationawareMotion2022}. Note, that considerations must be made for ill-posed motion planning problem; if the free-flyer moves with a high velocity, the time update for the position for the discrete system may jump quickly if the time interval is not fine enough. If the distance for two consecutive position states is larger than the smallest characteristic length of the ellipsoid, the algorithm will not detect a collision, although in reality (continuous time), a collision may occur. A simple method to mitigate this problem is to increase the sample resolution time and by interpolating the trajectory during the collision check \cite{jewison2017guidance}. 

\subsubsection{Ellipsoid-on-Ellipsoid Convex Collision Checking}
An improved collision-checking module, unusable for optimization-based planners except perhaps with pre-computed signed distance fields, uses any of a number of convex collision checking algorithms (e.g., \cite{yi-kingchoiContinuousCollisionDetection2009}) to compute convex-on-convex shape collisions. This allows, for instance, a bounding ellipsoid placed around robot collision geometry to be used without performing obstacle inflation for ellipsoid-on-point checking. Using an ellipsoidal-on-ellipsoidal representation, efficient convex collision checking algorithms may be employed for more precise three-dimensional collision detection; the approach of \cite{yi-kingchoiContinuousCollisionDetection2009} is applied for the real-time kino-RRT planner.

\subsection{Trajectory Smoothing by LQR Shortcutting}\label{sec:shortt}
Given the LQR-RRT* or other global trajectory which may contain jerky and unnatural paths, trajectory smoothing can be directly applied using a shortcutting approach which iteratively constructs path segments between existing nodes \cite{kallmann2008planning,geraerts2007creating,hauser2010fast}. This enables the generation of high-quality, collision-free, smooth paths that can be used directly for control. Specifically, LQR shortcutting algorithm is presented which generates dynamically feasible shortcut segments that are optimal to the linearized system and LQR cost function given by Eqs. \eqref{costlqr} and \eqref{immediateterminallqr}.
\begin{itemize}
\item \textit{Random Sampling}: From the initial trajectory $\sigma_0$, two nodes, $\mathbf{x}_a$ and $\mathbf{x}_b$, are randomly sampled which results in a trajectory with three sections, $\sigma_0=\left[\sigma_1,\sigma_2,\sigma_3\right]$. $\sigma_1$ is from the initial state to $\mathbf{x}_a$, $\sigma_2$ is from $\mathbf{x}_a$ to $\mathbf{x}_b$, and $\sigma_3$ is from $\mathbf{x}_b$ to the final state.
\item \textit{LQR Interpolation}: An interpolation using an LQR solution between $\mathbf{x}_a$ and $\mathbf{x}_b$ is determined. The new trajectory is $\sigma_{2,interp}$. This solution incorporates dynamic feasibility and optimality in the generated shortcut.
\item \textit{Collision Checking}: Lastly, the new trajectory $\sigma_{2,interp}$ is checked against any obstacles within the state-space which is discussed in the Collision Avoidance subsection. If the shortcut is collision-free, then $\sigma_{2,interp}$ is patched with the two other sections $\sigma_1$ and $\sigma_3$ into $\sigma_{new}=\left[\sigma_1,\sigma_{2,interp},\sigma_3\right]$.
\end{itemize}
By applying LQR shortcutting on the LQR-RRT* trajectory, high-quality, collision-free, smooth paths that are dynamically feasible for the on-orbit free-flyer is generated. Thus, control can be applied to track this trajectory.

\subsection{Information-Aware Motion Planning and Parameter Estimation}

Also known as active learning, the goal of information-aware planning is to enable the agent to actively explore the environment in order to improve accuracy and efficiency of the learning procedure. In an on-orbit assembly context, this would be useful when interacting with new payloads that may not be fully inertially characterized. In the case of parametric system identification procedures, information-aware planning occurs in the form of excitation trajectories, computed using constrained nonlinear optimization of metrics based on sensor information, also known as optimality criteria. An example of such a criterion is the minimization of the condition number, or the ratio between the largest and smallest singular values, of the input correlation matrix \cite{armstrong1989finding}. Some other criteria are based on the Fisher Information Matrix (FIM), which is a measure of the amount of information provided by the measurements about the unknown parameters. 

Consider the following discretized measurement model for the system described by \eqref{eq:bryce_dyn}, 

\begin{equation}
	\Tilde{\mathbf{y}}_k = h(\mathbf{x}_k,\mathbf{u}_k,\pmb{\theta}_k) + \mathbf{w}_{y, k}  
\end{equation}
where $\mathbf{w}_{y, k}$ represents the Gaussian measurement noise,  $\mathbf{w_y} \sim \mathcal{N}\left(0,\bm{\Sigma}_R\right)$,  and $\Tilde{\mathbf{y}}$, the measurements. The FIM, $\mathbf{F}$ is calculated as

\begin{equation} \label{eq:FIM}
\mathbf{F} = E \left[ \left[\frac{\partial}{\partial\bm{\theta}}\ln{[p(\Tilde{\mathbf{y}}|\bm{\theta}})] \right] \left[\frac{\partial}{\partial\bm{\theta}}\ln{[p(\Tilde{\mathbf{y}}|\bm{\theta}})] \right] ^T\right].
\end{equation}

where $E$ represents the expected value. If the information content in the measurement data is maximized, then it would also correspond to minimizing the lower bound on the variance of the estimates \cite{crassidis2004optimal}. 
Optimization of the FIM is therefore a good candidate for the design of information-rich trajectories. Measures of identifiability based on the information matrix, referred to as the alphabet optimalities \cite{siciliano2016springer} include A-optimality, which maximises the trace of the inverse FIM, or E-optimality, where the least eigen value of the FIM is maximized. Following previous research works by the authors \cite{albee2019combining,ekal2021online}, the A-optimality criterion is used for generating local information-aware motion plans in this work. 

For the on-orbit assembly scenario, performing a complete system identification procedure for each assembly component to be transported would require excessive expenditure of energy and time. Rather, the goal is to obtain a sufficient amount of uncertainty reduction post grasping, as the robot transports the component to the assembly location. This information, if taken into account, will enable more precise and optimal task execution without being overly cautious. The following discretized trajectory optimization problem is solved for obtaining on-the-fly adjustable information incentivization:
\begin{equation}
\label{eqn:opt}
    \begin{aligned}
        & \underset{\mathbf{u}}{\text{minimize}} & &J=
         \sum_{k = 0}^{N-1}{\mathbf{x}^\top_{t+k} \mathbf{Q} \mathbf{x}_{t+k} + \mathbf{u}^\top_{t+k} \mathbf{R} \mathbf{u}_{t+k}} + \mathbf{\Gamma}^\top \texttt{diag}\left(\mathbf{F}^{-1}\right)\\
         & \text{subject to}
         && \mathbf{x}_{t+k+1} = f(\mathbf{x}_{t+k},\mathbf{u}_{t+k}, \bm{\theta}), k = 0,\dots,N-1 \\
        &&& \mathbf{x}_{t+k} \in \mathcal{X}_{free}, k = 0,\dots,N,\\
        &&& \mathbf{u}_{t+k} \in \mathcal{U}, k = 0,\dots,N-1,\\
    \end{aligned}
\end{equation}
where $N$ is the horizon length and $\mathbf{Q}\succ0$ and $\mathbf{R} \succ 0$ are weighting matrices. The relative weighting term, $\mathbf{\Gamma} \in \mathbb{R}^j$, is used to tune the amount of information content of the trajectory. The output $\mathcal{P}_l := \{ \mathbf{x}_{t:t+N}, \mathbf{u}_{t:t+N-1} \}$ is made available for control over horizon $N$.

The evolution of $\mathbf{\Gamma}(t)$ can be designed based on dynamical and environmental considerations. A decay and eventual shut-off in the information gain content as the robot reaches the goal position, $\mathbf{\Gamma}$ can be modeled on heuristics such as a decreasing exponential. This work employs a decay method guided by the current covariance of the estimates \cite{albee2022rattle}, leading to $\mathbf{\Gamma}$ being zero once the current model characterization is within a desired precision tolerance. Selective information content can be assigned by tweaking weights individually per parameter. In other words, at local plan horizon starting at $t_{k+1}$, for all parameters of interest $i$ with variance $\sigma_{i, k}$, desired learning tolerance $\sigma_{n, i}$, the value of relative weighting per parameter, $\gamma_i$, is determined as shown in Fig. \ref{fig:algo}. Here, $\alpha$  determines the tolerance above a minimum desired parameter
variance, and $\beta$ is the decay constant.

\begin{figure}[!h]
  \begin{algorithmic}[1]
    \Procedure{\texttt{CovarWeight}}{$\mathbf{\Sigma_{n}}, \mathbf{\Gamma_0}, \alpha, \beta$}\
    \State $\mathbf{\Sigma}_{k+1} \gets \texttt{EstimationUpdateStep}(\mathbf{\Sigma}_k$)
    \For{$\forall i$}
      \If{$\sigma_i \leq \alpha \sigma_{n,i}$} \State $\gamma_i \gets 0$
      \Else \State $\gamma_i \gets \gamma_{i,0}e^{-\frac{\beta\sigma_{n,i}}{\sigma_i}}$
      \EndIf
    \EndFor
    \EndProcedure
  \end{algorithmic}
  \captionsetup{labelfont={bf}}
  \caption{The covariance-based information weighting procedure to determine parameter weightings can be automatically adjusted to within a learning tolerance.}
  \label{fig:algo}
\end{figure}

\subsubsection{Sequential Parameter Estimation} \label{sec:param_est}
Considering that tracking current model knowledge is of interest, a rapid and sequential method of estimation, such as recursive least squares is used for inertial parameter estimation. The estimation problem can be formulated as
\begin{equation}\label{eq:estim_formulation}
    \tilde{y} = \mathbf{H}\theta  + v,  \quad \quad v\sim N(0,\mathbf{W})
\end{equation}
Where $\theta = \left[\begin{array}{cccc}
1/m  & 1/I_{xx} & 1/I_{yy} & 1/I_{zz}  
\end{array}\right]^T$ are the mass and moments of inertia parameters respectively. $\tilde{y}$ represent the measurement data, in this case linear acceleration, $\bm{a}$ and angular acceleration, $\bm{\alpha}$ (obtained from differentiation of angular velocities). Representing the control inputs as $\mathbf{F} \in \mathbb{R}^3 $(forces)  and $\bm{\tau} \in \mathbb{R}^3$ (torques),  the regressor matrix $\mathbf{H}$ can be written as
\begin{equation}
    \mathbf{H} = \left[\begin{array}{cc}
         \mathbf{F} & \mathbf{0}_{3\times 3}\\
         \mathbf{0}_{3\times 1} & \bm{\tau} \mathbf{I}_{3\times3}
    \end{array}\right]
\end{equation}
Resembling the update step of the Kalman filter without a process model, the recursive least squares algorithm at every time step is given as \cite{crassidis2004optimal}
    \begin{gather*} \label{eq:est_update}
        \theta_{k+1} = \theta_k  + \mathbf{K}_k (\tilde{y}_k - \mathbf{H}_k\theta_k)\\
        \mathbf{P}_{k+1} = (\mathbf{I} - \mathbf{K}_k\mathbf{H}_k)\mathbf{P}_k
\end{gather*}
where 
    \begin{gather*}
    \mathbf{K}_k = \mathbf{P}_k\mathbf{H}_k'\left(\mathbf{H}_k\mathbf{P}_k\mathbf{H}'_k +\mathbf{W} \right)^{-1}
\end{gather*}
Details on outlier rejection and accounting for latency are discussed in \ref{sec:improving}.


\subsection{Robust Model Predictive Control}
A control framework providing robustness guarantees against unstructured uncertainties throughout the system, $\mathbf{w}_x \in \mathbb{W}$ from \eqref{eq:traj_opt}, is needed to track the trajectories planned by Section \ref{sec:plan}. Ideally, this approach should also be flexible in its modeling capability and capable of handling real-time computation. Robust tube model predictive control fits these criteria, making it an appropriate choice for trajectory tracking of local plan output, requiring a few ingredients: a reference trajectory $\mathbf{x}_{ref}(i)$, a reference input $\mathbf{u}_{ref}(i)$, a dynamics model $\mathcal{M}$, and a predicted bound on disturbances to the discrete-time dynamics to form robustness guarantees. $\mathbf{w}_x$ supplies the uncertainty, and is often conveniently modeled as a set of box constraints,
\begin{align}
	\mathbb{W} = \left\{ ^I\mathbf{w} \in \mathbb{R}^n : \begin{bmatrix}
	\mathbf{I}_6 \\
	-\mathbf{I}_6 
	\end{bmatrix} \mathbf{w} \leq
	\begin{bmatrix}
	^I\mathbf{w}_{max}\\
	^I\mathbf{w}_{max}
	\end{bmatrix}
	\right\}.
\end{align}

For the linear time-invariant case (e.g., translational dynamics), the free-flyer error dynamics are then

\begin{align}
\mathbf{{x^+}}_{err} = \mathbf{A}\mathbf{x}_{err} + \mathbf{B}\mathbf{u}_{err} + \mathbf{w}_x.
\label{eqn:err}
\end{align}

where $\mathbf{x}_{err} = \mathbf{x} - \mathbf{x}_{ref,i}$ and $\mathbf{u}_{err} = \mathbf{u} - \mathbf{u}_{ref,i}$. $\mathbf{w}_x$ can be thought of as a combination of any unstructred noise uncertainty, as well as any uncertainty due to imperfect system modeling (e.g., incorrect parameter estimates, $\hat{\bm{\theta}}$.

The stochastic dynamics are now in a suitable format for linear robust tube MPC. First, a disturbance rejection controller called the ancillary controller must be used to reject disturbance from a nominal MPC trajectory. The ancillary controller takes the form below, and is added onto a nominal MPC input, $\mathbf{v}$:
\begin{align}
	\mathbf{u}_{anc} &= \mathbf{K_{anc}}(\mathbf{x}-\bar{\mathbf{z}})\\
	\mathbf{u} &= \mathbf{v} + \mathbf{u}_{anc}
	\label{eqn:control}
\end{align}
where $\mathbf{v}$ is a nominal actuation determined by a deterministic MPC and $\mathbf{K}_{anc}$ indicates the ancillary controller disturbance rejection gain, and $\bar{\mathbf{z}}$ is a nominal state not necessarily the same as the real initial state. $\mathbf{K}_{anc}$ can be determined through a simple LQR procedure on the nominal dynamics, but provides optimal performance and better robustness guarantees when determined via a tube minimization procedure \cite{Buckner2018a}. That is, if the robust positively invariant set (RPI) can be minimzed through the choice of $\mathbf{K}_{anc}$ then more rigorous guarantees exist; namely, the tube robustness guarantee that if the system state $\mathbf{x}$ starts within a set $\mathbb{Z}$ centered around a planned control trajectory $\mathbf{z}$, under the given uncertainty and ancillary controller it will remain with a tube around this trajectory for all possible $\mathbf{w}_x$ disturbances. The tube robustness guarantee can be thought of simply as ``if you start in the tube, you stay in the tube."

Secondarily, the actual nominal MPC trajectory $\mathbf{z}$ used by the ancillary controller must be determined. The error dynamics of eq. \ref{eqn:err} are used.
$\mathbb{U}$ and $\mathbb{X}$ are converted to tightened constraints, effectively giving up control authority to the ancillary controller for disturbance rejection. These tightened constraints are indicated as $\mathbf{x} \in \bar{\mathbb{X}} \subset \mathbb{X}$ and $\mathbf{u} \in \bar{\mathbb{U}} \subset \mathbb{U}$ and are derived from the nominal box constraints, $\mathbb{X}$ and $\mathbb{U}$. The exact constraint tightening procedure can be found in both Buckner and Limon \cite{Buckner2018a} \cite{limonDesignRobustTubebased2008}. A notable feature observed in this constraint tightening procedure is the fact that large uncertainty bounds will make constraint tightening infeasible. This serves as a notification that the considered uncertainty levels are beyond the system's actuation and/or dynamics capability to adequately counter. It is possible to compromise and settle for a lower level of robustness with a lower assumed $\mathbf{w}_x$ or to even require replanning at prior stages of the pipeline. Finally, the nominal MPC (which does not necessarily align with the initial real state, $\mathbf{x}_i$) can be found by solving

\begin{align*}
	\underset{\mathbf{u}(i),\bar{\mathbf{z}}_0}{\text{min}}\ & \quad J= \sum_{i=0}^{N-1}[\bar{\mathbf{z}}_i-\mathbf{x}_{i,ref}]^\top\mathbf{Q}[\bar{\mathbf{x}}_i-\mathbf{x}_{i,ref}] + [\bar{\mathbf{v}}_i-\mathbf{u}_{i,ref}]^\top\mathbf{R}[\bar{\mathbf{v}}_i-\mathbf{u}_{i,ref}] +\\
	&\hphantom{JJ} [\bar{\mathbf{z}}(N)-\mathbf{x}_{ref}(N)]^\top\mathbf{H}[\bar{\mathbf{z}}(N)-\mathbf{x}_{des}(N)]
	\\
	\text{subject to}&                                                                                                                                                                      \\
	&\quad \mathbf{z}_{i}^+ = f(\mathbf{z}_{i},\mathbf{v}_{i}),                                                                                                                                   \\
	&\quad \bar{\mathbf{z}} \in \bar{\mathbb{X}}\\                                                                                                                          
	&\quad \bar{\mathbf{v}} \in \bar{\mathbb{U}}\\ 
	&\quad \bar{\mathbf{v}} \in \mathbf{x}\bigoplus(-\mathbb{Z}).\\ 
\end{align*}

Where $\bigoplus$ represents the Minkowski sum. Eq. \eqref{eqn:control} is executed for the first timestep of the nominal MPC solution until the solution can be recomputed at $i^+$.

A key observation explored in recent work by Albee and Ekal \cite{albee2022rattle} is that reachable sets can sometimes be updated in real-time based on available data. Re-computation of $\mathbb{Z}$, then, might allow one to ease off of overly conservative robustness guarantees based on the latest available system data (e.g., learned inertial parameters).

\begin{figure}[htb!]
    \centering
	\includegraphics[width=0.55\linewidth]{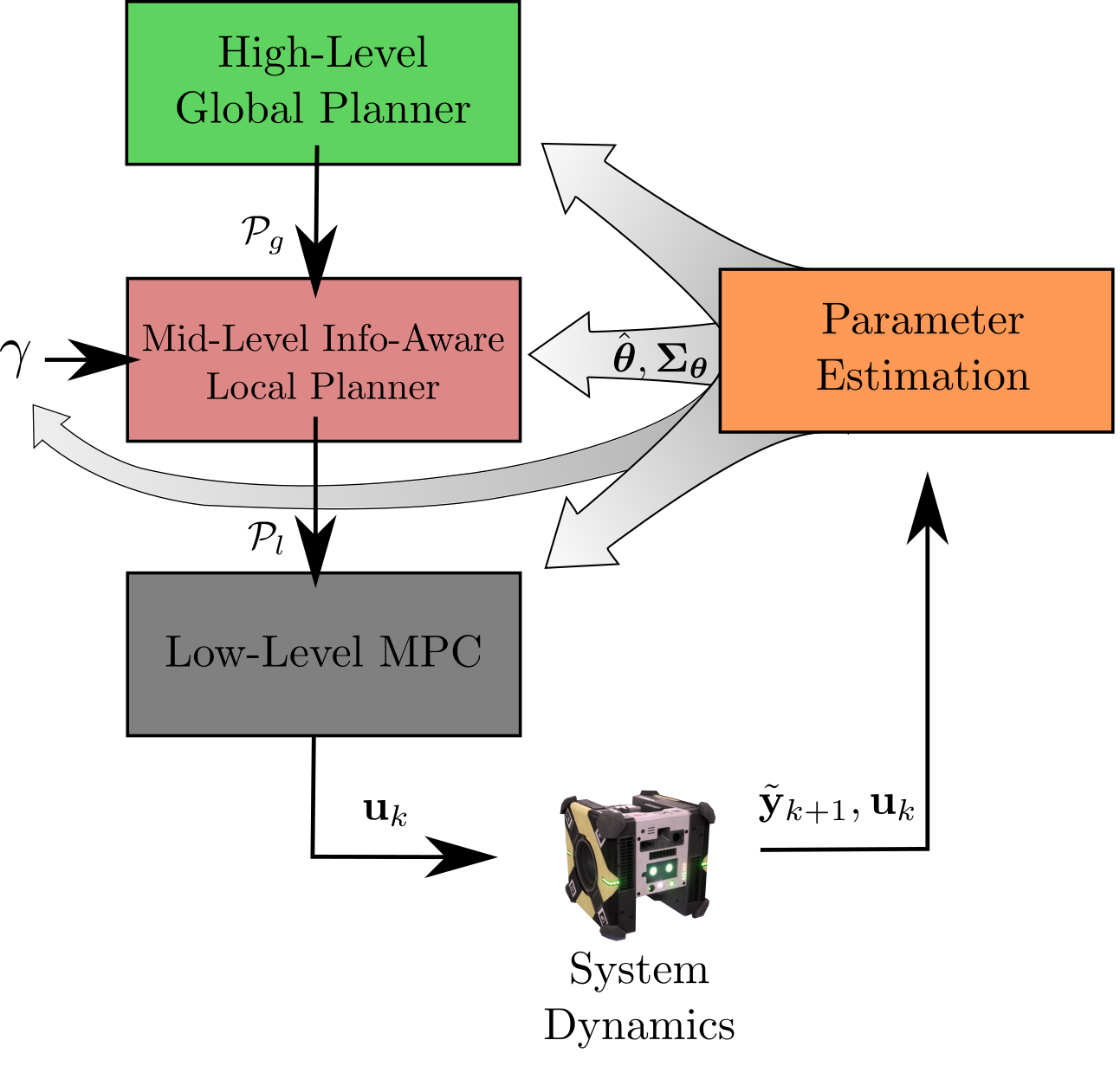}
   	\captionsetup{labelfont={bf}}
	\caption{A sketch demonstrating components of the RATTLE planning framework, a high-level (global) long-horizon kino-RRT planner, a mid-level, shorter-horizon planner incorporating information awareness via an adjustable weighting term, $\gamma$, and an online update-able robust MPC. \cite{albee2022rattle}}
	\label{fig:summary}
\end{figure}
\subsection{The RATTLE Motion Planning Algorithm} The RATTLE algorithm combines many of the planning, estimation and control components described above. It has a multi-tiered framework consisting of a global planner (Kino-RRT), a local, receding-horizon information-aware planner and a low-level, robust model predictive controller (Fig. \ref{fig:summary}). An accompanying parameter estimator provides the latest estimates of the system model. While the algorithm performs online parametric uncertainty reduction through exploratory motions, it also provides robustness guarantees against current model uncertainties and environmental disturbances. Further, the local planner and controller possess online update-able models, which means that information about latest model knowledge is used to inform subsequent planning and control actions. The global planner, if capable of online replanning, facilitates the accommodation of dynamic obstacles in the environment. The result of RATTLE is a reactive, robust approach for dealing with parametric uncertainty during task execution.

\subsection{The Combined Approach: Planning, Control, and Estimation Techniques for Free-Flyer On-Orbit Assembly} \label{sec:approach}
These planning and control components can be assembled in two primary modes of operation: (1) offline, pre-motion computation and (2) online, real-time computation. Offline computation provides benefits in terms of additional time to refine global plan solutions, create robustness guarantees, and other computationally-intensive operations. Offline computation might be desirable if on-orbit operations will permit additional time in creating motion plans, if model certainty is high, or if refined motion plans are highly desirable. Meanwhile, online computation is desirable if models or environmental constraints are unknown or changing, or if additional computation time is unacceptable for the operation of interest. At the cost of producing solutions that might be sub-optimal, online computation allows adaptation to changing conditions and replanning capability.

These offline and online computation methods are combined in a complete on-orbit assembly scenario, which consists of three types of maneuvers:

\begin{enumerate}
    \item Offline planning that can be produced in waiting areas near the construction site. System model knowledge is considered adequate and robust tracking is unnecessary.
    \item Online planning, control, and estimation that is responsive to poor model knowledge and has a time sensitivity---real-time plans are desirable.
    \item Offline planning that can be produced in passive waiting areas near the safe area. System model knowledge is considered adequate, but there is uncertainty that makes robust tracking desirable.
\end{enumerate}

\begin{figure}[hbtp!]
    \centering
	\includegraphics[width=0.6\linewidth]{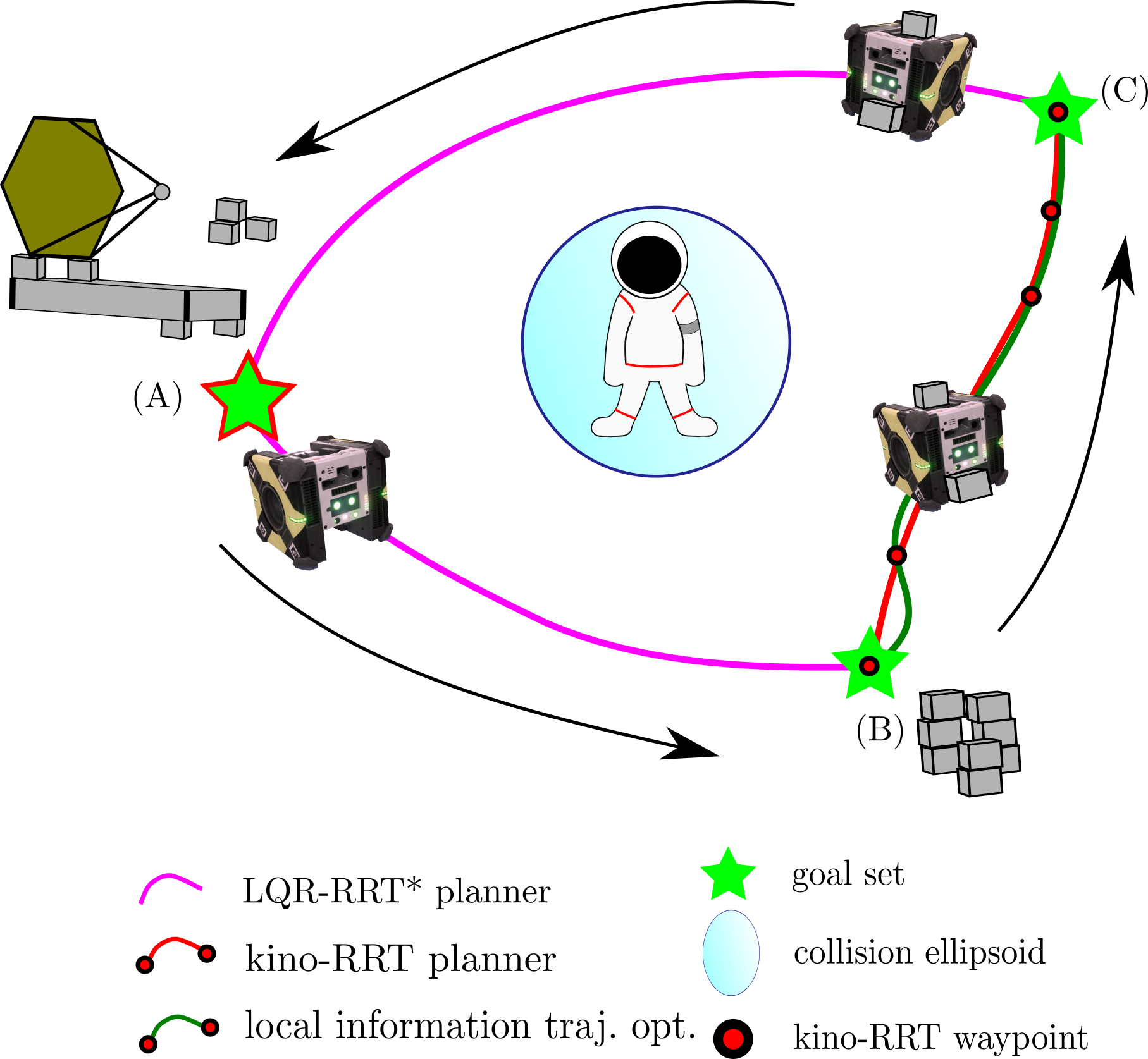}
   	\captionsetup{labelfont={bf}}
	\caption{The on-orbit assembly sequence, combining elements of online and offline planning, robust control, and model improvement.}
	\label{fig:test13}
\end{figure}

These maneuvers are demonstrated between three waypoints, (a), (b), and (c). These waypoints represent an assembly ``construction site area", a ``storage area" and a ``safe area," respectively and are illustrated in Fig. \ref{fig:test13}. At waypoint (b), the storage area, the free-flyer acquires new payload to transport, modifying its model inertial parameters, $\bm{\theta}$.  The (a)-(b) maneuver is an offline plan without robust tracking; the (b)-(c) maneuver uses the RATTLE framework to perform model improvement post payload capture; and the (c)-(a) maneuver is an offline plan with robust tracking. Combined, these maneuvers represent typical situations that a free-flyer manipulating payload might find itself in, with respect to the safety of its current surroundings, the quality of its model knowledge, and the time urgency of progressing toward its goal. All maneuvering takes into account a central obstacle constraint (posed as ellipsoid-on-point for LQR-RRT* and ellipsoid-on-ellipsoid for kino-RRT).

\begin{table}[hbtp!]
    \centering
    \caption{Planning and control elements for each portion of the on-orbit assembly sequence. Components indicated are for the maneuver immediately following the listed waypoint.}
    \begin{tabular}{|c|c|c|c|c|}
        \hline
        Waypoints & Designation & Online Planning & Robust Tracking & Model Improvement  \\
        \hline
        \hline 
         A-B & Construction Site & \xmark & \xmark & \xmark\\ 
         B-C & Storage Area  & \cmark & \cmark & \cmark \\ 
         C-A & Safe Area & \xmark & \cmark & \xmark \\ 
         \hline
    \end{tabular}
    \label{tab:my_label}
\end{table}

\section{Implementation, Environment, and Operational Details}
\label{sec:setup}
\begin{figure}
    \centering
    \includegraphics[width=0.65\linewidth]{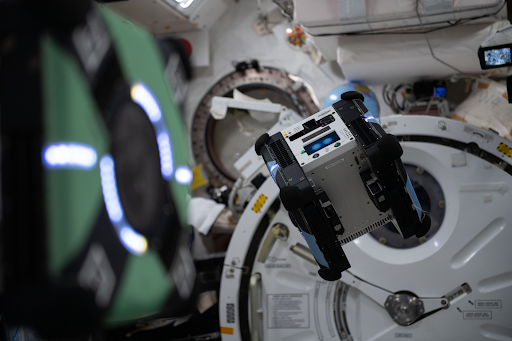}
    \caption{The Astrobee robots operating aboard the ISS. (Credit: NASA)}
    \label{fig:astrobee}
\end{figure}

The on-orbit assembly modules of Section \ref{sec:methods} were implemented to run on the robot alongside the Astrobee Flight Software \cite{bualat2015astrobee,smith2016astrobee} aboard the International Space Station (ISS). The Astrobee robots, shown in Fig. \ref{fig:astrobee}, are cube-shaped 30 cm wide programmable robots on the ISS that have been reconfigured by the ReSWARM experiments for autonomy and controls research \cite{Albee2020guide}. This section overviews some of Astrobee's key software and environmental details, as well as some of the unique obstacles that arose in configuring the robots for on-orbit assembly experimental testing.

\subsection{Key Software Information}
The Astrobee Robot Software uses ROS as middleware for communication, with about 46 nodelets grouped into approximately 14 processes running on two ARM processors \cite{Fluckiger} (with one additional Android-based High-Level Processor). The Astrobee Robot Software also includes a high-fidelity simulator, which enables testing of developed algorithms before implementation on hardware; in simulation, a single local computer simulates running all nodes/nodelets as if both of Astrobee's main processors are available. The physics simulator is essentially a set of plug-ins for the Gazebo robot physics simulator, and offer the same ROS interfaces for sensor communications as the hardware. The ROS/Gazebo-based simulation environment includes extensive modeling of Astrobee including its impeller propulsion system, onboard visual navigation, environmental disturbances, and other high-fidelity models \cite{Fluckiger}.

Astrobee's default flight software (FSW) consists of planning (Planner), mapping (Mapper), factor-graph based localizer (called the EKF module due to a historical version of the localizer), control (CTL), and force application (FAM) modules. Coordinating these components is a managing node, the choreographer. These elements are shown in Fig. \ref{fig:fsw}. These key autonomy interfaces are overviewed next.

\subsubsection{Localization}
Astrobee's localization takes in measurements from various sensors, $\tilde{\mathbf{y}}$ and fuses these together in conjunction with a measurement and dynamical model, $h(\mathbf{x})$ and $f(\mathbf{x}, \mathbf{u})$, to produce a state estimate $\mathbf{\hat{x}}$. The measurement model might be very complex or impossible to model in the case of vision-based estimation systems, for example. In these cases, sophisticated algorithms produce pose/state estimate information which is fused alongside other sensor information, like IMU data. Astrobee, for example, has three main measurement sources:

\begin{itemize}
  \item Visual-inertial odometry (VIO)
  \item Sparse mapping
  \item IMU data
\end{itemize}

Errors in any one of these measurement sources could imperil the state estimation algorithm. For Astrobee, the IMU is particularly noisy, VIO is sensitive to fast attitude movements (motion blur), and sparse mapping relies on an accurate, recent map with clear visual features. (A common issue depending on the recency of mapping activities is the quality and accuracy of the map, which might be distorted and out-of-date.\footnote{
There are a few methods to determine if localization data is accurate in the absence of ground truth:

\begin{itemize}
  \item Get ground truth---sometimes sources like video footage can give clues if localization is obviously inaccurate.
  \item Check for physically infeasible changes in state. Compare estimates against the largest deviations possible given the system dynamics. (This might also be used as a check to reject bad estimates.)
  \item Look at estimation signs of health, like feature counts, that show if measurements might be troubled.
  \item Check for small-scale ``jumpiness'' that might lead to controller problems.
\end{itemize}
})

Astrobee's localization system was overhauled in early 2021 to use a new graph-based system utilizing GTSAM, named AstroLoc \cite{soussan2022astroloc}. AstroLoc has higher accuracy that filtering-based methods, is fast enough for real-time use, and handles cheirality issues. It is heavily VIO-reliant since landmarks frequently change (i.e., camera image changes are carefully analyzed to estimate motion, combined with IMU info). The latest iteration of the localization system runs at about $5\ [\text{Hz}]$, though updates upsample to a rate of $62.5\ [\text{Hz}]$ (half the IMU update rate).

\subsubsection{Controller}
The controller, after receiving state estimates from the localization system, takes a desired trajectory $\mathbf{x}_{des}$ and tuning parameters, ultimately producing a requested actuation, $\mathbf{u}$. By default, Astrobee uses a PD controller for the attitude and position controller running at $62.5\ [\text{Hz}]$. Astrobee's controller does not account for actuation saturations.

\subsubsection{Mixer}\label{sec:mixer}
Astrobee has a holonomic thruster placement, with 12 independent thrust vents. It generates thrust by drawing in air through two central impellers and expelling it through the vents. The mixer takes requested inputs $\mathbf{u}$ and maps these to permissible actuator commands, i.e., vent opening angles. A thruster placement diagram and explanation is given in \cite{Smith2016}. The physical properties of the impeller and the nozzles, such as the nozzles' minimum/maximum open angles, etc. can be found in \cite{Albee2020guide}, from which the mixer can be determined. For Astrobee, the mixer is not as simple as the typical matrix found in impulsive systems like the SPHERES robots \cite{Jewison2014}. Actual forces and torques applied may be significantly different from those requested by the control if the control is physically infeasible (i.e., does not satisfy $\mathcal{U}$), leading to control input saturation.

\subsection{Environments}
The Astrobee simulation testing environment includes both a granite table world for simulating 3 DoF planar ground testing and a space station environment for simulating full 6 DoF testing. Testing in simulation and multiple subsequent ground test opportunities enabled ReSWARM's transition to hardware and the evaluation of its performance on real resource-constrained processors leading up to microgravity testing.

\begin{figure}[h!]
  \centering
  \includegraphics[height=6.5cm]{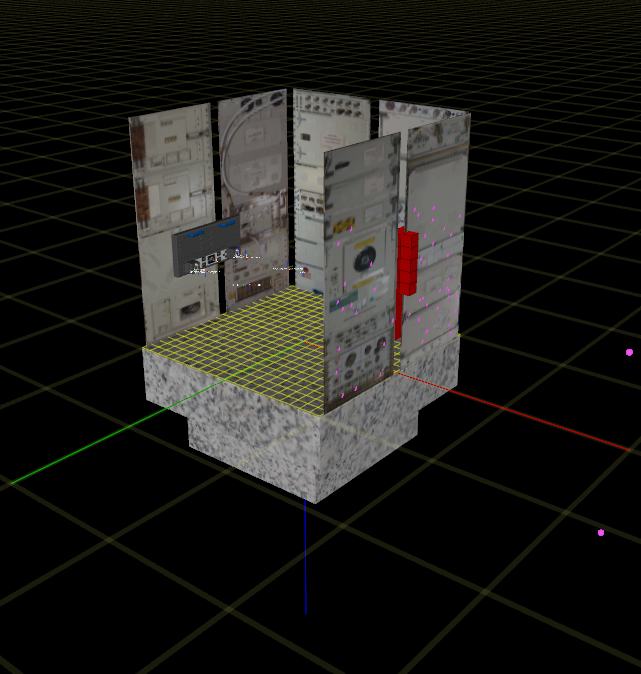}
  ~~~~
  \includegraphics[height=6.5cm]{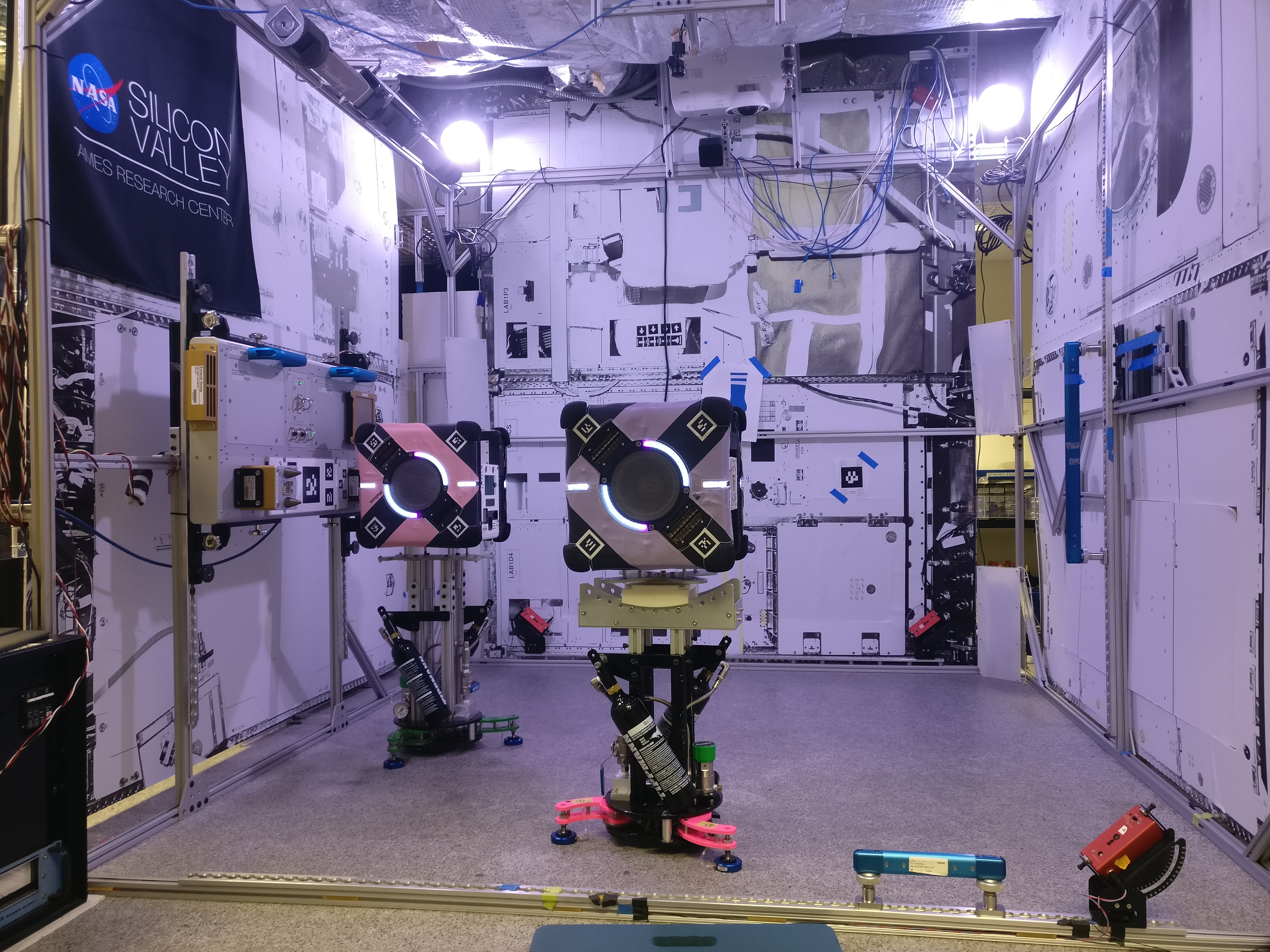}
  \caption[Astrobee's simulation ground environment.]{Astrobee's simulation and physical granite environment. The coordinate convention for granite world (left). RGB corresponds to XYZ for the axes shown, where z+ points down. The granite table ground facility at NASA Ames (right), with Astrobees Bsharp and Wannabee shown.}
  \label{fig:sim_axes}
\end{figure}

\subsubsection{Ground (Granite) Environment}
The granite world in simulation mimics the granite table in use at the NASA Ames ground test facility, where the robots are placed on air-bearing surrounded by a mock-up of the interior of the space station . The simulation table is $2 \times 2\ \text{[m]}$. The coordinate system mimics that of the ISS, where z+ is toward GND (down). The simulation and physical ground facility are shown in Fig. \ref{fig:sim_axes}.

\subsubsection{ISS Environment}
\begin{figure}[h!]
  \begin{center}
    \includegraphics[height=6cm]{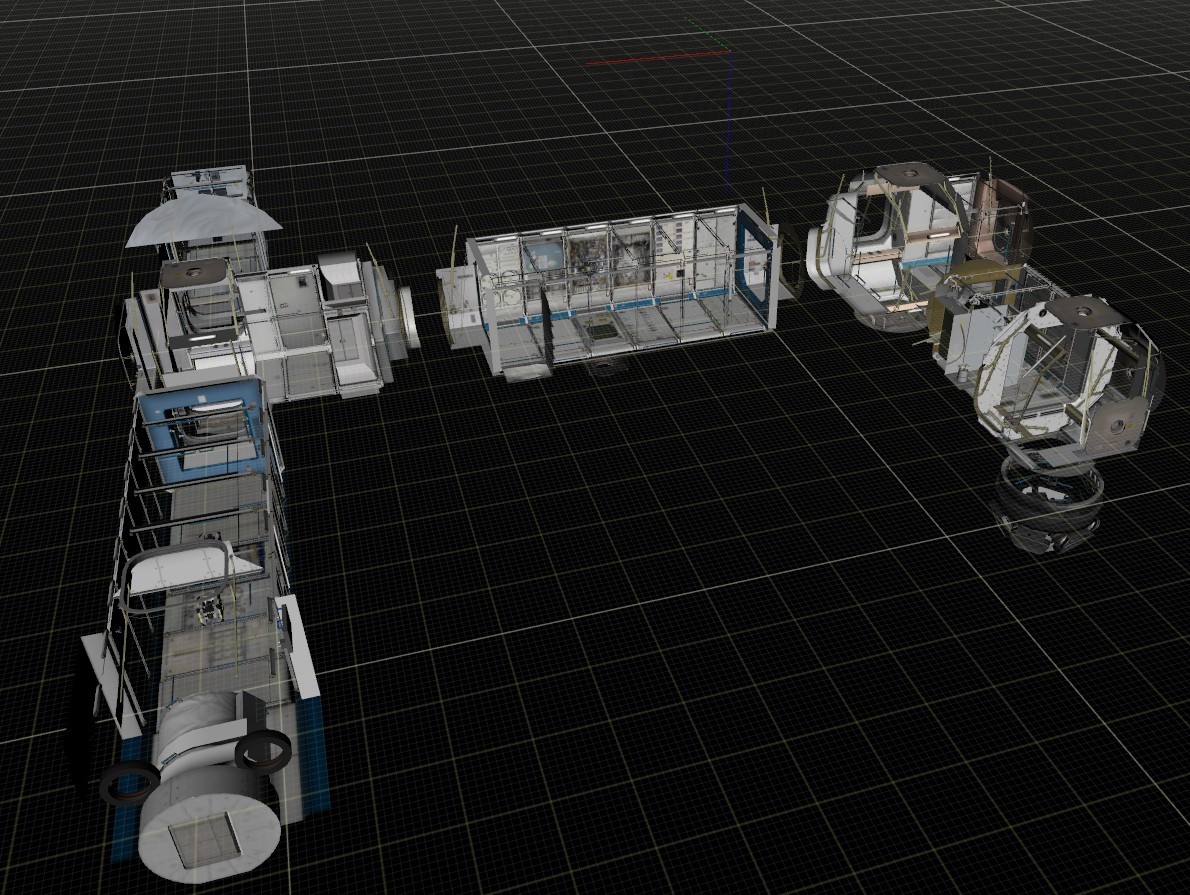}
    \includegraphics[height=6cm]{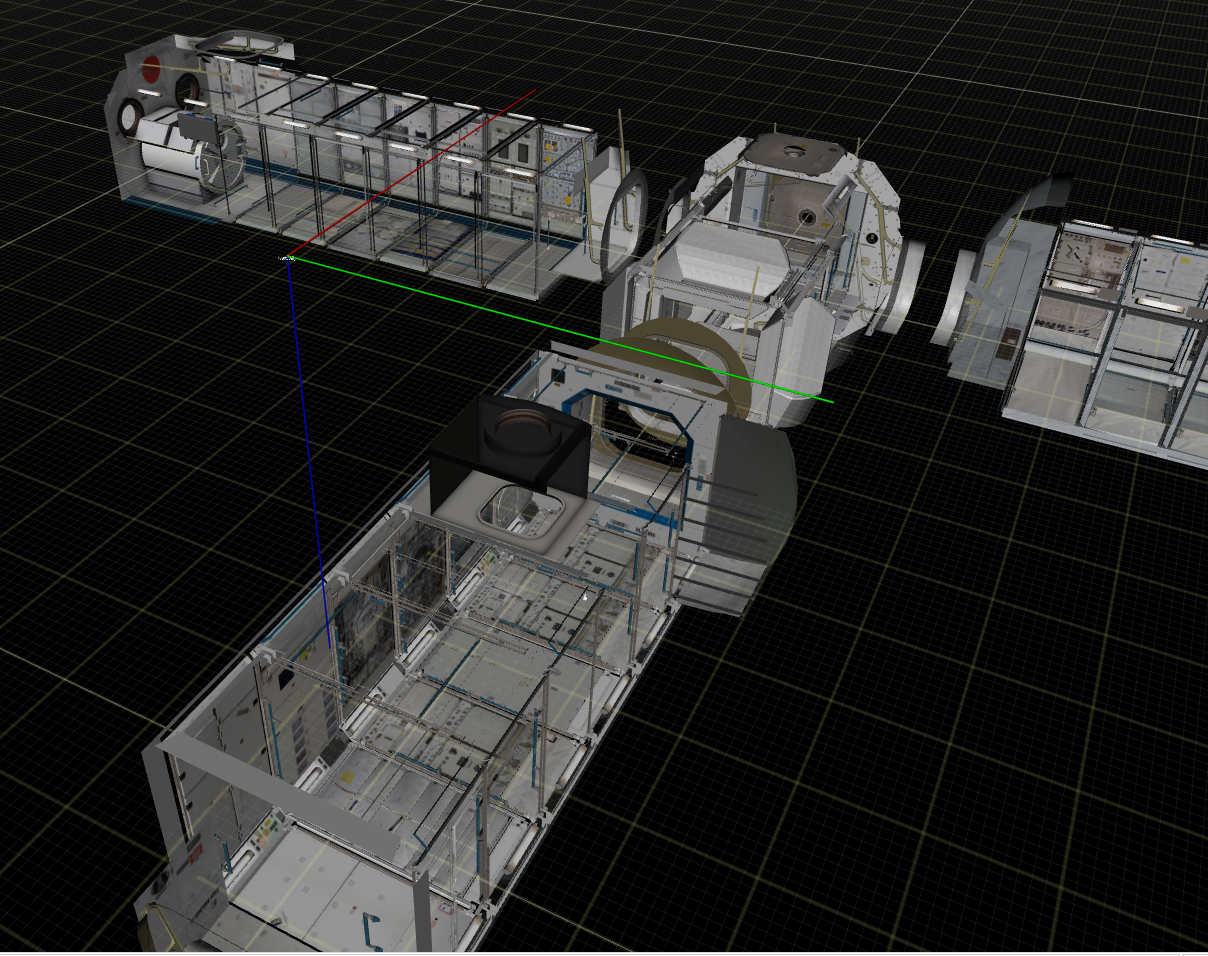}
  \end{center}
  \caption[The US Orital Segment.]{The US segment of the International Space Station, within which Astrobee is permitted to operate. At right, RGB axis colors corresponds to XYZ.}
  \label{fig:iss_env}
\end{figure}

\begin{figure}[h!]
  \centering
  \includegraphics[width=0.8\textwidth]{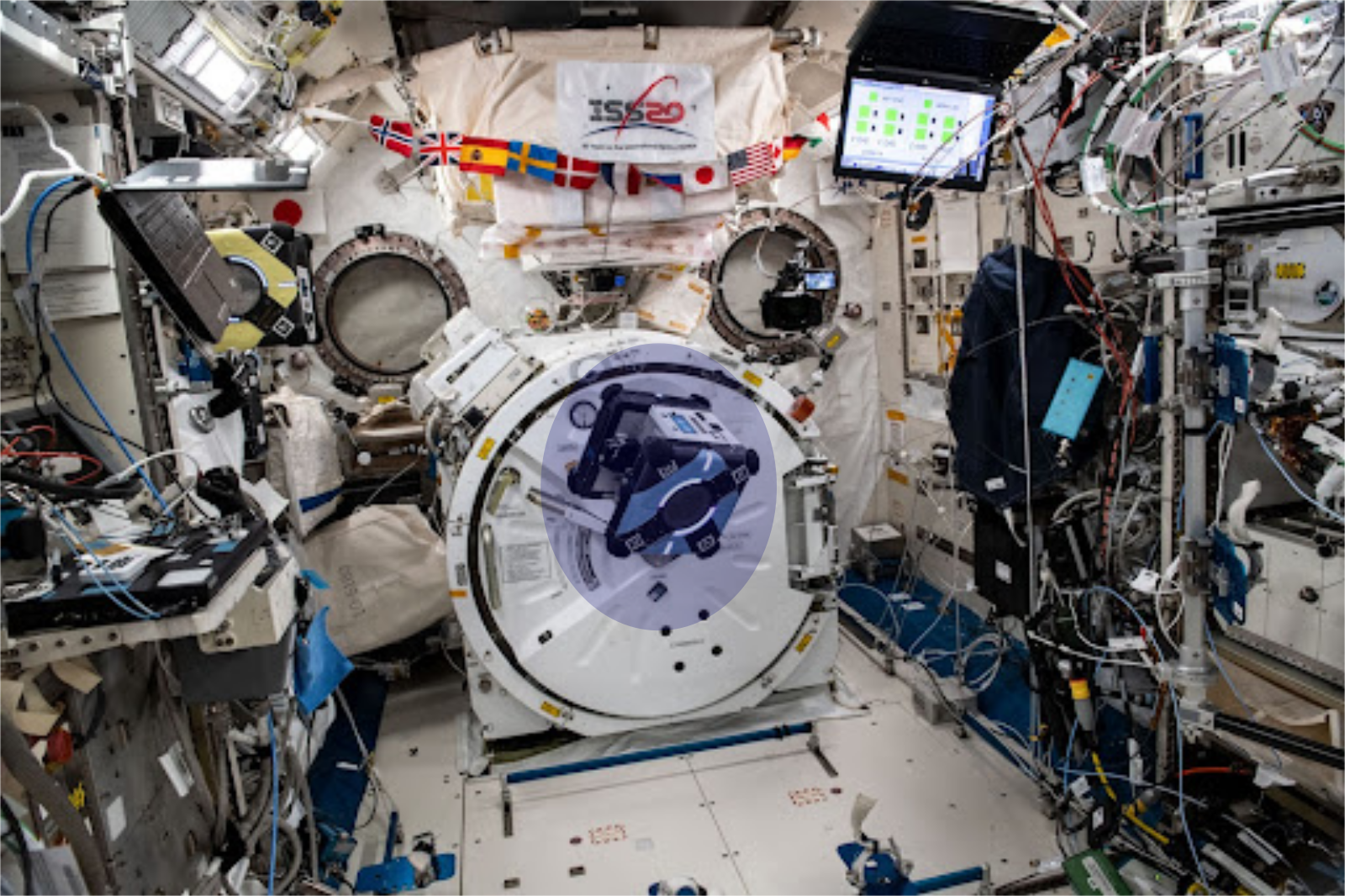}
  \caption[The JEM on ISS.]{The JEM on the ISS which serves as the on-orbit test facility. Astrobee Bumble is highlighted in the center. This particular view is looking PORT (in the +y-axis direction).}
  \label{fig:ISS_hardware}
\end{figure}

The ISS simulation world is a mockup of the US Orbital Segment of the International Space Station, shown in Fig \ref{fig:iss_env}. As in the ISS facility, Astrobee is docked in the Japanese Experiment Module (JEM) of this segment and operates primarily within that module. The approximate volume of this segment is $1.5 \times 6.4 \times 1.7\ \text{[m]}$, in ISS coordinates. The coordinate system and simulation environment (with JEM in the upper left) is shown in Fig. \ref{fig:iss_env}. The actual ISS facility is shown in Fig. \ref{fig:ISS_hardware}.

\subsection{Actuation and Motion Constraints}
Astrobee has limits on the geometry it can safely traverse, as well as maximum forces and torques that its mixer system can generate. These constraints affect the planning and control, and are documented here.

\subsubsection{Force and Torque Constraints}
The maximum thrusts per axis at various impeller fan speeds are given in Table \ref{table:speeds}. Vents on the x-axis have the largest nozzles and thus have the maximum acceleration capability, as the table shows. These maximum force values are approximate based on empirical data produced by NASA Ames. Astrobee's thruster offset is about $0.1\ \text{m}$ from the center of mass, so the torque limit is very roughly about $\frac{1}{10}$ of each of these values. 

\begin{table}[h!]
  \centering
  \caption{The approximate thruster maximum forces per axis.}
  \begin{tabular}{ |c|c|c|c| } 
    \hline
    Motor Speed [RPM]& x-axis [N]& y-axis [N]& z-axis [N] \\
    \hline
    \hline
    2000 RPM & 0.452 & 0.216 & 0.257 \\
    \hline
    2500 RPM & 0.680 & 0.332 & 0.394 \\
    \hline
    2800 RPM & 0.849 & 0.406 & 0.486 \\
    \hline
  \end{tabular}
  \label{table:speeds}
\end{table}

\noindent

To mitigate the reduction in localization accuracy due to loss of features, the motion planning algorithm used more conservative limits than those presented in Table \ref{table:speeds}. 

\subsubsection{Position and Velocity Constraints}
\begin{figure}
    \centering
	\includegraphics[width=0.6\linewidth]{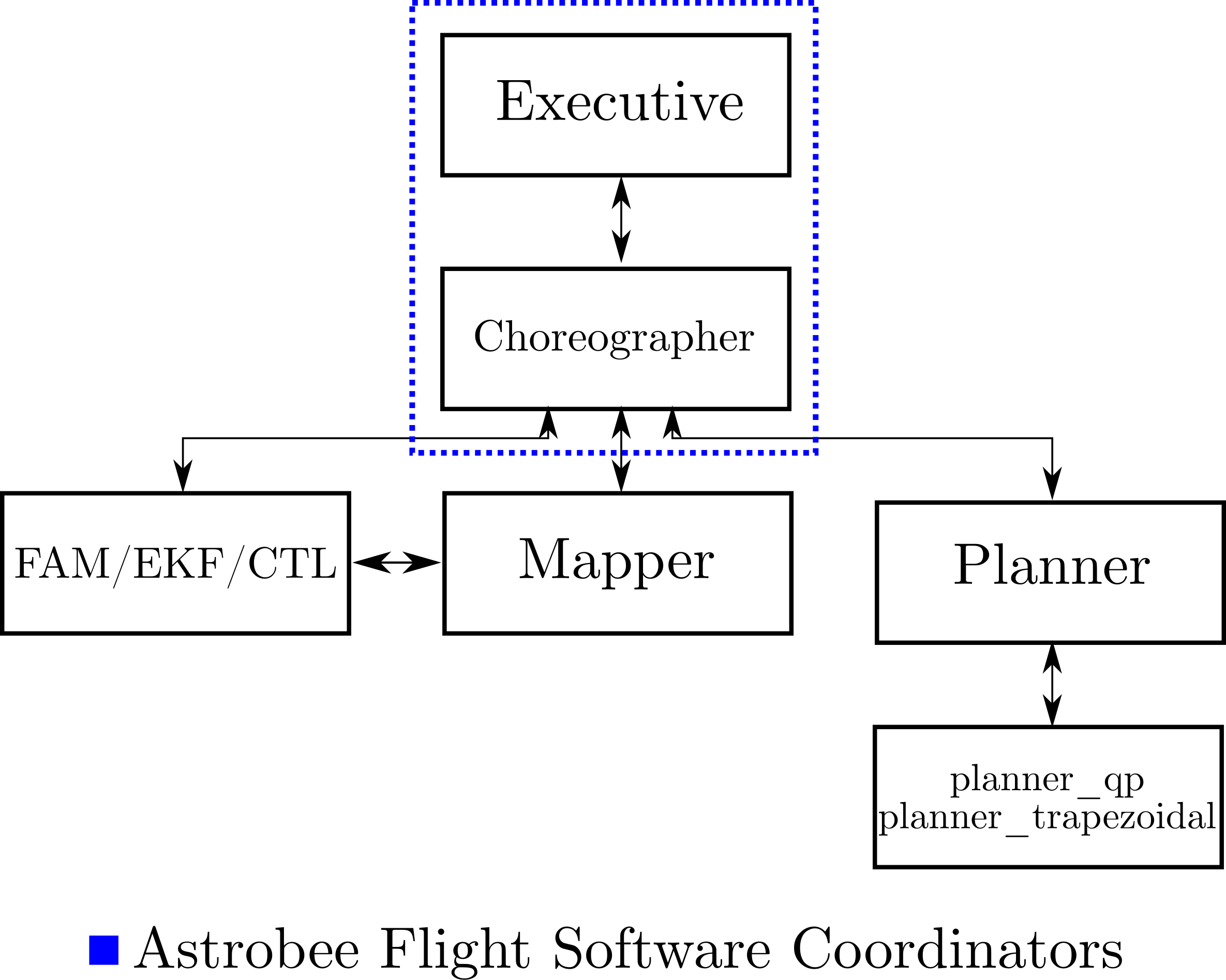}
   	\captionsetup{labelfont={bf}}
	\caption{A high-level overview of the relationship of some the Astrobee FSW's default autonomy modules.}
	\label{fig:fsw}
\end{figure}

\mbox{}\\
Astrobee has a number of keep-out zones defining rough exterior boundaries of the enclosing environment. Position constraints are not necessarily enforced by any default Astrobee planner or controller (one planner, \texttt{planner\_qp}, see Fig. \ref{fig:fsw}, does obey keep-in/keep-out zones.).  In practice, position constraints must be estimated based on actual setup conditions, especially so for the ISS. On ISS, cargo and objects are regularly reconfigured. The authors have found the following constraints to be a good ``safe'' approximation of the JEM interior volume, in coordinates relative to the approximate JEM centroid located at $(10.9, -6.65, 4.9)\ [\text{m}]$,

\begin{align}
\begin{split}
-0.65 \leq &x \leq 0.65\\
-3.2 \leq &y \leq 3.2\\
-0.8 \leq &z \leq -0.8\ [\text{m}].
\end{split}
\end{align}

\begin{figure}[h!]
  \begin{center}
    \includegraphics[width=0.5\textwidth]{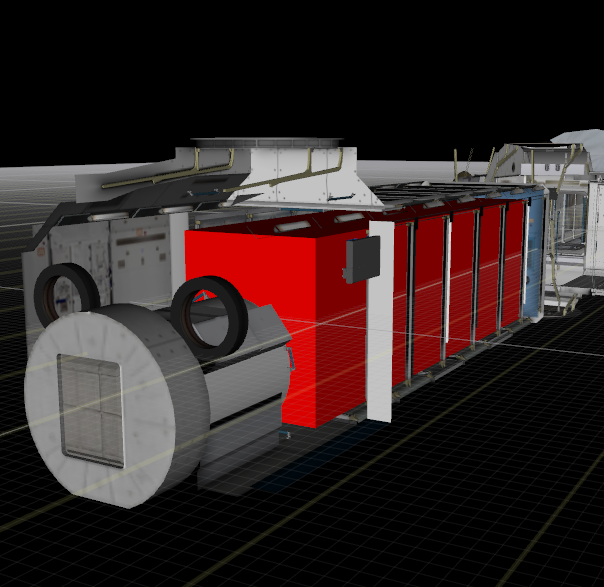}
  \end{center}
  \caption{The approximate usable volume of the JEM, in red.}
  \label{fig:iss_jem}
\end{figure}

Approximate keep-in zones are shown in Fig. \ref{fig:iss_jem}. By default, Astrobee uses a $\pm 0.1 \ \left[\frac{\text{m}}{\text{s}}\right]$ velocity constraint for safety reasons.

\subsection{Special Considerations for Hardware Deployment}\label{sec:considerations}
In addition to algorithmic details, some implementation novelties were required for the successful use of the on-orbit assembly modules within the Astrobee software stack, showing some of the important considerations when moving to hardware.

\subsubsection{Software Architecture and Complexity}
The middleware coordination of the architecture is non-trivial, and deserves careful consideration to ensure model updates, planner outputs, and other components are on-time and properly shared. On hardware, this role is actually a software package unto itself. While the low-level details of implementing this coordination are not covered here, it is important to ensure that all modules are working from the same set of data and are receiving updates (such as model or constraint updates) at the same time.

\begin{figure}
    \centering
	\includegraphics[width=\linewidth]{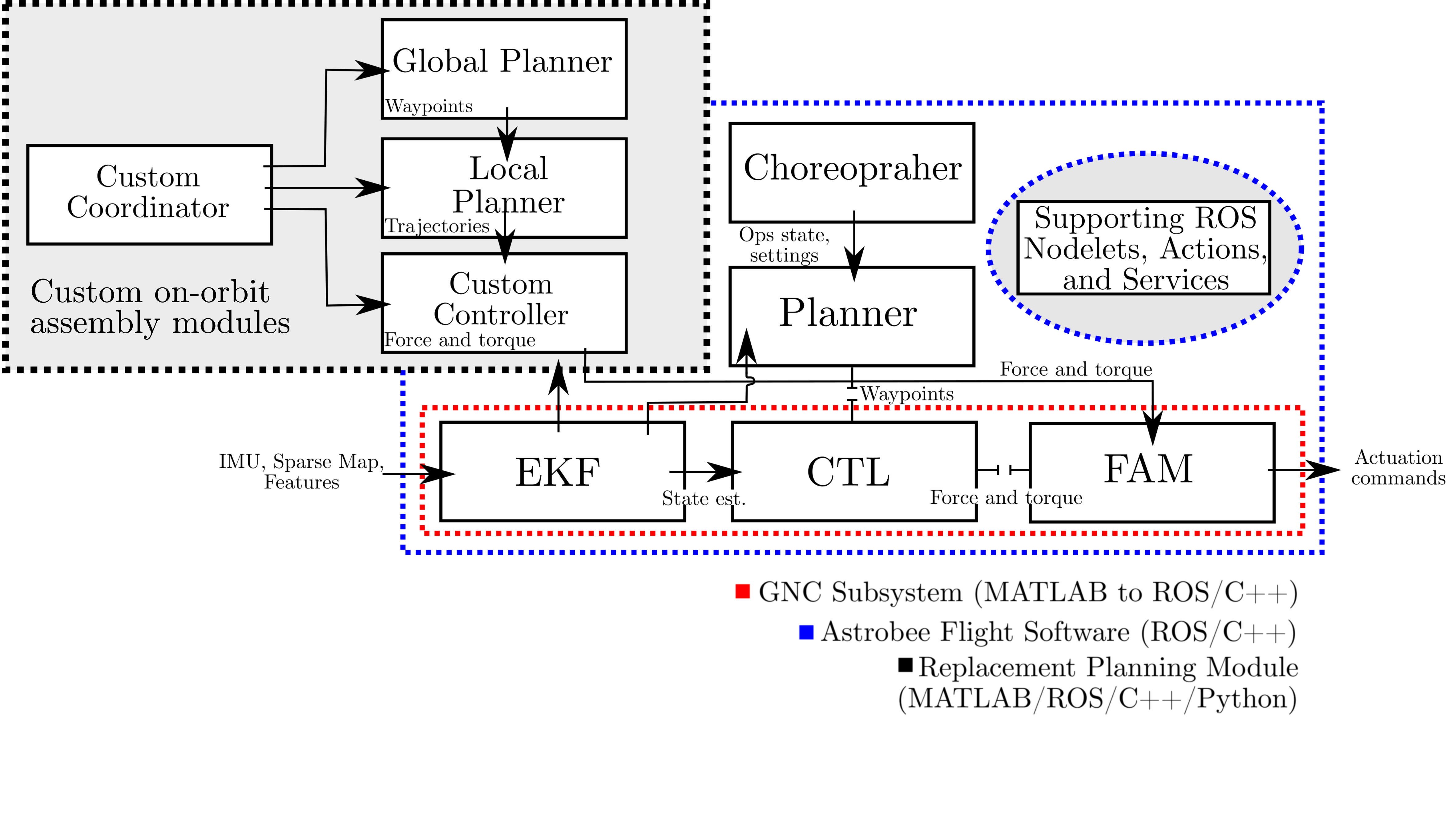}
   	\captionsetup{labelfont={bf}}
	\caption{Additions to Astrobee's default flight software, including a custom planning and control pipeline. An approximation of the additional nodes implemeted for ReSWARM testing are shown in gray.}
	\label{fig:fsw_override}
\end{figure}

Getting code onto the Astrobee hardware in a safe and reliable way was one of the chief concerns of algorithm integration. The components of Astrobee's default software stack needed to be overridden to run custom autonomy modules. This is possible by adding entirely new nodes that publish ROS messages on the same topics that low-level control elements might expect and taking care not to trigger the default software stack. This is important because the default software stack can be left entirely untouched and additional nodes may be added as extra packages on top of existing flight software packages. This greatly simplifies integration and verification. An example of this process is shown in Fig. \ref{fig:fsw_override}. Additionally, standardization such as working from a single operating system, development environment, and middleware framework (ROS) helps tremendously with this task.

Multiple external libraries, integrated within the Astrobee flight software stack, were used for on-orbit integration. The ACADO toolkit \cite{houskaACADOToolkitAnOpensource2011} was used for solving nonlinear programming of the local planner for parameter learning, and CasADi \cite{anderssonCasADiSoftwareFramework2019} for implementation of a robust tube MPC. Additionally, Bullet Physics' C++ collision checker \cite{coumansBullet83Physics} and Autograd were used for collision detection and additional automatic differentiation, respectively. Cross-compilation of these libraries proved a significant challenge for hardware integration and should be carefully noted when moving to hardware deployment. Highlights of the Astrobee software stack and other implementation hurdles are discussed further in \cite{Albee2020guide}.

\subsubsection{Importance of Localization}
Astrobee's default localization system is under active development, and requires special care in map-building and motion constraints to avoid severe pose estimate jumps, a major challenge for control and inertia estimation. This caused issues for experiments using RATTLE that encouraged rotational motion --- in particular information-seeking motion for inertia estimation --- sometimes causing significant ``jumps'' in localization estimates. Localization is the building block of the rest of the autonomy stack, and so localization issues are troubling for all other components. In some cases, this might be viewed as an additional aleatoric disturbance source on the system dynamics. Localization required special consideration for ISS testing which is discussed in Section \ref{sec:flightconsider}.

\subsubsection{Timing Considerations, Planning Rates}\label{sec:ratess}
The various modules run at different rates to accommodate processor capabilities; in addition, elements of Astrobee's default flight software have their own rates, characteristics, and computational loads. Getting the timing right for the various online components of an autonomous system like Astrobee is key for successful operation (especially when the software middleware does not have any real-time guarantees and is simply best-effort---real-time in the sense of a real-time operating system). For instance, Astrobee reports its applied forces and torques (wrench) after passing through the mixer at approximately $62.5$ [Hz]. However, this rate can decrease based on processor load. Additionally, reported wrenches are often out-of-order, requiring careful chronological reordering to enable parameter estimation. This is discussed further in Section \ref{sec:improving}.

\begin{figure}[hbtp!]
  \centering
  \includegraphics[width=0.9\linewidth]{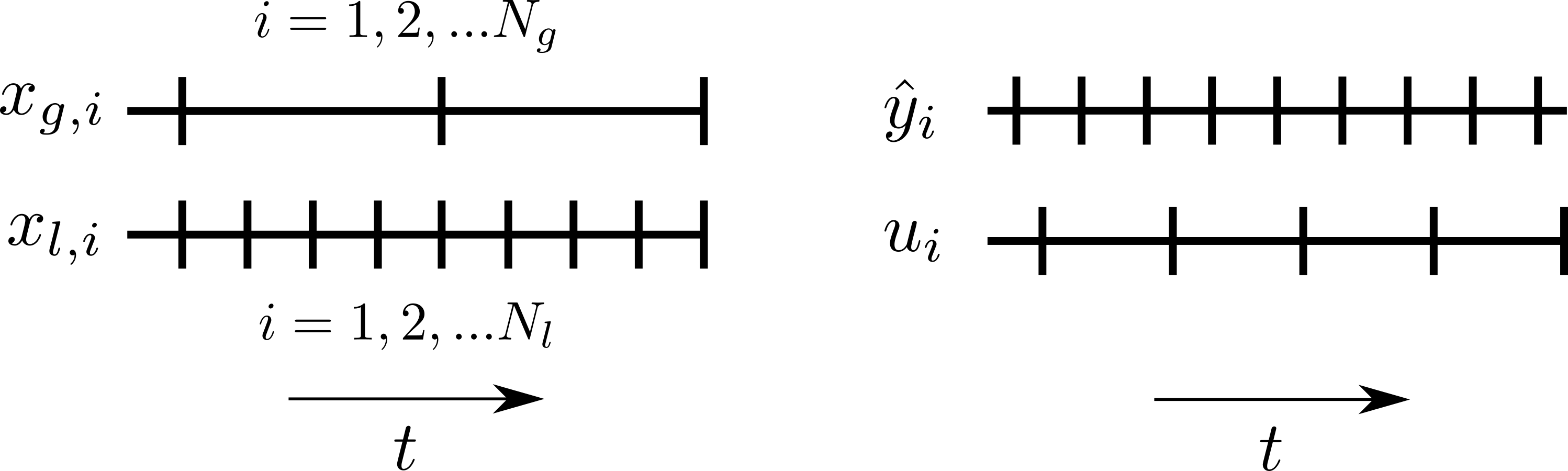}
  \caption[Syncing data rates on hardware.]{Syncing rates for planning, estimation, and other modules is very important to avoid getting inaccurate data. Here $N_g$ and $N_l$ denote global and local planning horizons, while $x_{g,i}$ and $x_{l,i}$ denote global and local plan states; $\hat{y}_i$ and $u_i$ denote estimated states and inputs, respectively.}
  \label{fig:rates}
\end{figure}

A simple version of the problem is demonstrated in Fig. \ref{fig:rates}. On the left plot, the global and local plan have different trajectory frequencies. The global plan waypoints, which might be selected from a higher-fidelity global plan, should coincide with the fixed local planner horizon, which operates over an $N_l \cdot t_l$ length time horizon. It is possible to use differing rates for both planners, but then interpolation will be required. On the right plot of Fig. \ref{fig:rates}, the localization output and reported input do not match up at all, and one data stream might even have variable rates. Again, computations involving the discrete dynamics rely on syncing up function inputs at the same time---a propagation of the model dynamics for an input one second too late could be catastrophic, for instance. In this case, one would need to decide what discretization time to use for the system models, and again might rely on some form of interpolation scheme (or use variable-length integration of the continuous dynamics) to achieve desired results for e.g., the parameter estimation module.

If interpolation is required, special care must be taken for interpolation in non-Euclidean manifolds. The quaternions, $\mathcal{H}$, are parameterized in $\mathcal{SO}(3)$. Simple linear interpolation of Euler angles yields very poor approximations of interpolation along the the 4D unit sphere. Two relatively simple methods provide good quaternion linear and great circle interpolation: \texttt{LERP} and \texttt{SLERP} \cite{shoemake1985animating}. $\texttt{SLERP}$ results in manifold-accurate interpolation, with constant angular velocity.

Finally, the frequencies at which the global planner, local planner, controller, and other components may run at are limited by length of the computation time required on the hardware of the desired robot. Balancing computational burden to obtain sufficient speed on hardware is critical. ISS hardware rates are provided later, in Table \ref{tab:comp_speed}.

During the extensive ground testing, the runtimes for the RRT$^*$ planners, local planners, MPC horizons, and rate of actuation commands were extensively fine-tuned to ensure real-time operation and planning on desired 6 DOF motion and rates of the Astrobee robots. The overall goal for each Astrobee test was to have each run last less than 5 minutes for the most complex scenario, to maximize extremely precious on-orbit test time. It was also internally noted by previous Astrobee experiments prior to ReSWARM that the ISS Astrobees had more running processes in the background compared to the ground Astrobees. Thus, it was expected that the computational speeds for planning and control were to decrease moving from the ground hardware to the ISS hardware. In consideration of this knowledge, the sample rate for e.g., control was set to 5 [Hz], and the controller defined was adjusted to this rate accordingly to reduce the time complexity of the planning problem and meet the performance requirement.  

    
    
    

\subsection{Special Considerations for Operations}
On-orbit ReSWARM operations required special considerations for preflight, flight, and postflight procedures to mitigate any anomalies during the mission. Field testing using the Astrobee robots is extremely precious time, and therefore preparation and rehearsal for flight activities was key. Some of this special planning is noted here.

\subsubsection{Preflight}
Preflight operations were performed at the NASA Ames' granite table facility and the Astrobee flight software was used for software integration. Preflight integration occurred in two major ways:

\begin{enumerate}
  \item The simulation work was integrated with the Astrobee flight software using the ROS/Gazebo environment. The team developed a number of interfaces to enable the on-orbit testing using Astrobee's default flight software \cite{Albee2020guide}.
  \item The Astrobee flight software was integrated with the hardware and tested at the Ames granite table facility. Integration was performed over multiple day-long ground test sessions for algorithm improvements and hardware bug fixing. Ames personnel assisted with much of the in-person testing.
\end{enumerate}

\subsubsection{Flight}\label{sec:flightconsider}
Flight operations were conducted over a months-long process of developing ground and crew procedures to ensure rehearsed readiness for operations day. The crew procedure provides instructions of how ISS crew members should aid in experimentation which follows the general format summarized below:

\begin{enumerate}
  \item Crew member sets up Astrobee(s) in their home position.
  \item Astrobee(s) hold their position, and crew member releases the Astrobee(s) and moves to a waiting area out of the field of view (FOV).
  \item Astrobee(s) begin execution of a set of maneuvers designed to demonstrate an aspect of the algorithms of interest.
  \item Test concludes, or it is stopped early by the ground operator.
  \item Crew grabs Astrobee(s), and a new test is selected. Repeat the procedure for additional tests.
\end{enumerate}

Additionally, operations required several roles to make the sessions run smoothly. Operations personnel adhered to the ground procedures which provided instructions on setting-up the Astrobees, setting up the Ground Data System (GDS), running the tests, troubleshooting the Astrobees, and wrapping-up the session. The roles required for operations are summarized below.
\begin{itemize}
  \item CLI OP: Command-Line interface operator. The person sending command-line commands to the robots, responsible for real-time debugging, and applying workarounds.
  \item GDS OP: The person sending GDS commands to the robot and responsible for nominal test commanding. 
  \item POIC: The person at the Payload Operations Integration Center responsible for communicating with the crew 
  \item CREW: The crew member on the ISS assisting with experiments and responsible for setting up and monitoring tests.
  \item MIT: The ground crew at MIT.
  \item AMES: The ground crew at NASA Ames.
\end{itemize}

ISS operations were conducted with the help of a single flight crew member, whose role was to observe experiments and place two Astrobee robots at their respective start positions at the onset of testing. For ReSWARM-1 and ReSWARM-2, Astronauts K. Megan McArthur and Matthias Maurer assisted in the respective Astrobee sessions. Experiments were conducted in a set of short (2-5 minute)  tests that demonstrated various aspects of algorithm of interest. The flight crew required a crew procedure ahead of the test session in order for the session's planning to be clearly communicated in a single checklist-type summary document.

For ReSWARM-1, MIT and AMES worked together remotely in which MIT commanded the GDS OP and CLI OP roles while AMES commanded the POIC role. For ReSWARM-2, MIT and AMES worked onsite together at NASA Ames with MIT commanding the GDS OP and CLI OP roles. 

A major flight challenge was the performance of the Astrobee's default localization. Astrobee's localization relies on identifying features within a known map of the ISS, which is a challenging task for a constantly-changing interior. For ReSWARM, tests were designed that minimized rotational motion and pointed toward feature-rich areas to assist in localization performance. When Astrobees were in an initial feature-poor area, crew were provided guidance (following procedure protocol) to initially move Astrobee to feature-rich areas and back to the initial position to lock onto its localization.

\subsubsection{Postflight}
The postflight operations included the delivery of ISS data files, imagery, and video from the experiments. Software as carefully configured to enable accurate, annotated, data recording for all desired test sessions. Downlink of this data was performed by NASA Ames, allowing reconstruction of the important on-orbit information detailing algorithm performance, discussed in the next section.

\subsubsection{Anomalies}
Due to the hardware anomalies that occurred on the Honey Astrobee in the hours before ReSWARM-1, ReSWARM-1 consisted of single-Astrobee operations with 28 tests. Afterwards, the Honey Astrobee was downmassed, and the Queen Astrobee was commissioned to bring back the multi-Astrobee capability on the ISS. This delayed ReSWARM-2 by a few months, but with more time to update algorithms. A total of 39 successful test runs (an informal Astrobee test record) including multi-Astrobee tests which belonging to a separate investigation in ReSWARM-2 were later demonstrated.

\section{Results: Microgravity Testing for On-Orbit Assembly}\label{sec:results}
The three-step on-orbit assembly sample scenario, using the methods described in Section \ref{sec:approach} was executed on the JEM onboard the ISS. Results presented in the following sections are divided based on the key tools used for executing the on-orbit assembly sequence given in Fig. \ref{fig:test13}: The LQR-RRT* global planner for offline global planning in the presence of obstacles; the RATTLE information-aware motion planning framework with model improvement for learnable parametric uncertainty; and tube MPC for robust tracking in the presence of disturbances. Section \ref{sec:ratess} on planning and control speeds addresses real-time execution of these components on hardware, discussing computation times, time horizons, and other implementation details.   

\subsection{Experiment Set 1: LQR-RRT* Global Planning}

\begin{figure}[hbtp!]
  \centering
  \includegraphics[width=1.0\linewidth]{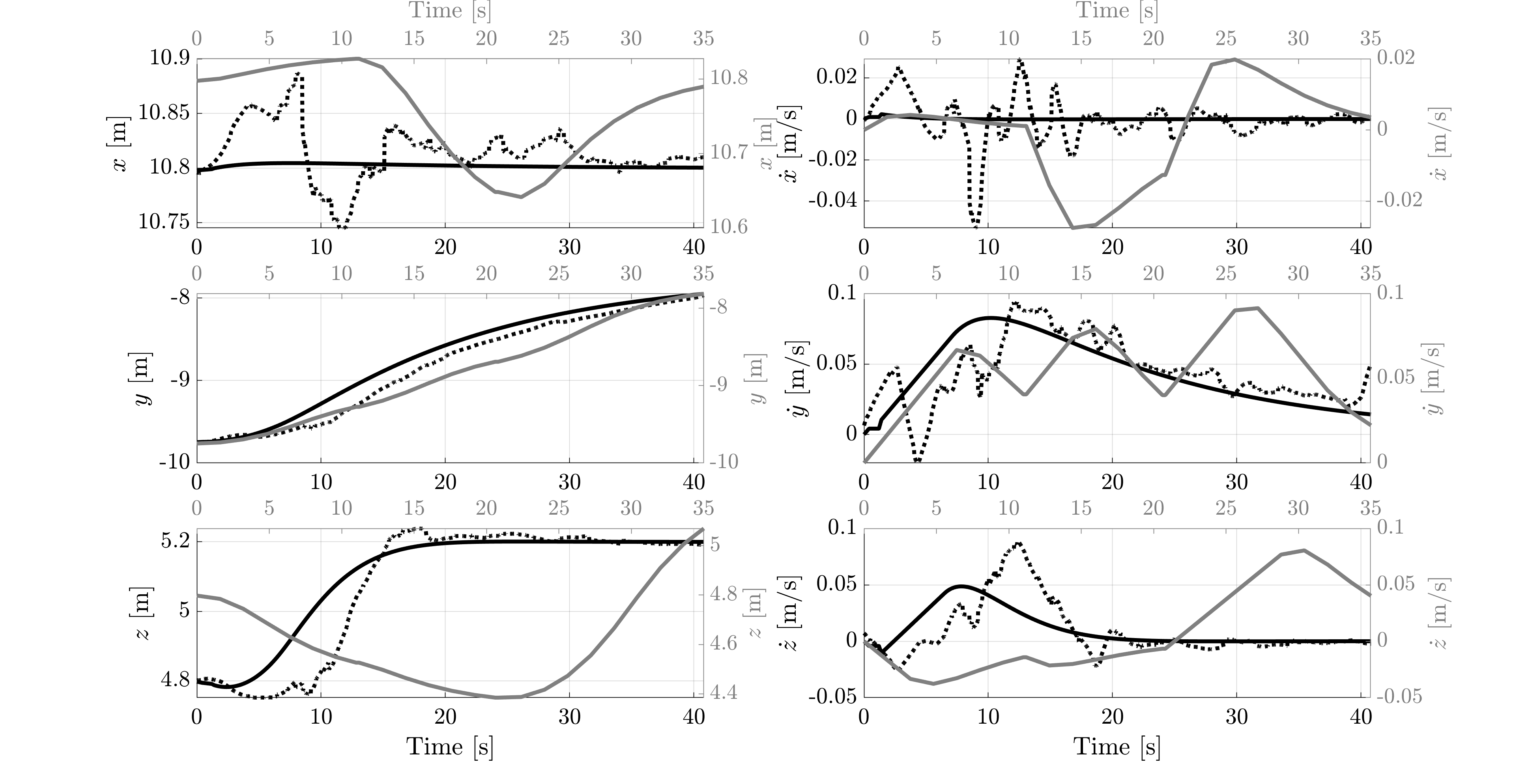}
  \caption[Tracking the MIT reference trajectory.]{Translational trajectories for the Astrobee robot using LQR-RRT* (solid gray) and LQR shortcutting (solid black), and response  from robust tube MPC (dotted black).}
  \label{fig:Exp1xyz}
\end{figure}
\begin{figure}[hbtp!]
  \centering
  \includegraphics[width=1.0\linewidth]{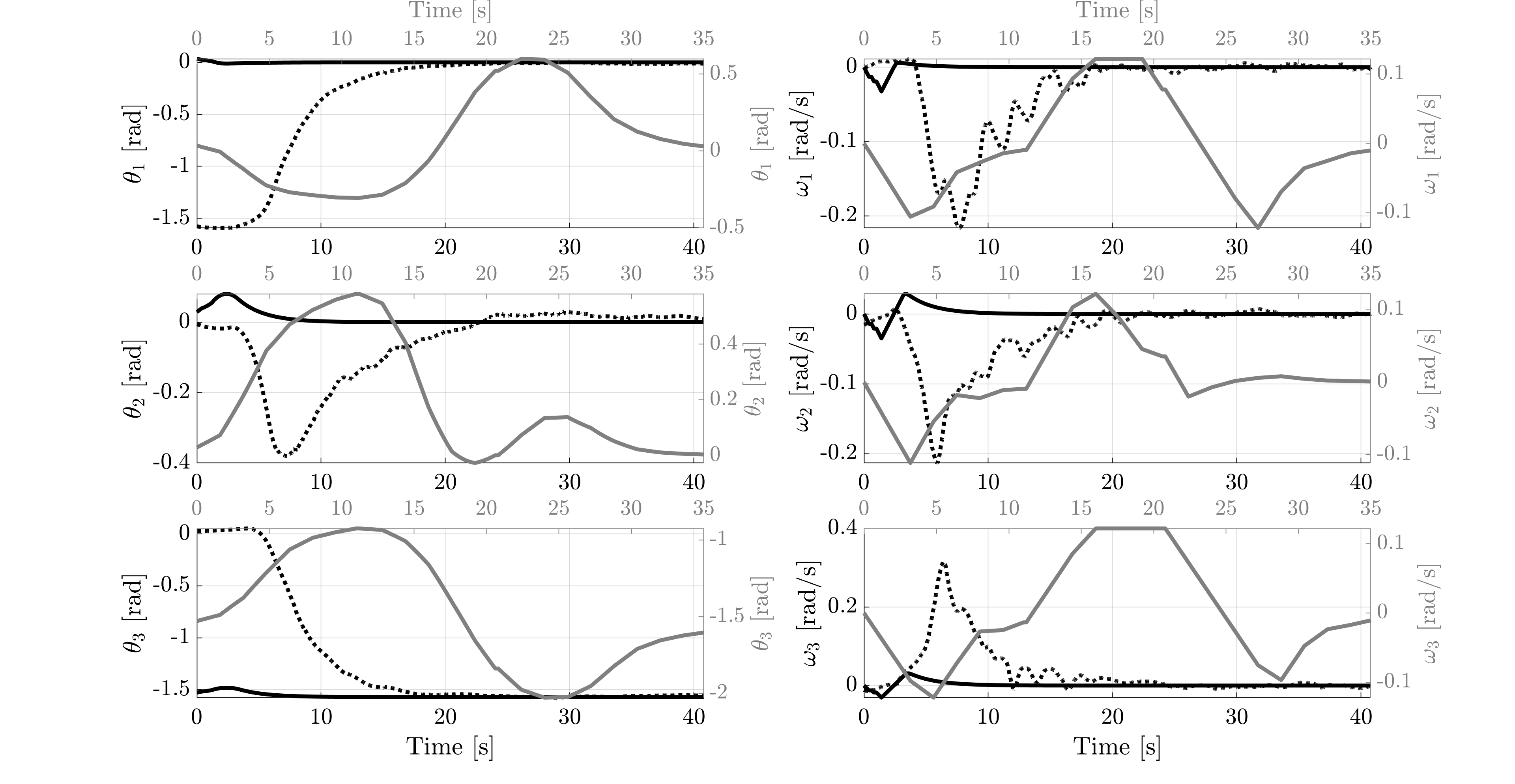}
  \caption[Tracking the MIT reference trajectory.]{Attitude trajectories for the Astrobee robot using LQR-RRT* (solid gray) and LQR shortcutting (solid black), and response  from robust tube MPC (dotted black).}
  \label{fig:Exp1theta}
\end{figure}

In this initial experiment, the Astrobee robot is tasked with moving between two points in the presence of obstacles. The obstacles present for this experiment include 6 safety ellipsoids which constrain the robot inside the test environment. Additionally, a static obstacle is generated inside the test environment. Trajectories are planned using LQR-RRT* and LQR shortcutting, and the planned trajectory is tracked using robust tube MPC to counteract disturbances. Figures \ref{fig:Exp1xyz} and \ref{fig:Exp1theta} show the trajectories obtained from LQR-RRT* (solid gray line) and LQR shortcutting (solid black line), and the response from robust tube MPC (dotted black line). At specific points along the path, the LQR-RRT* trajectory is jerky and unnatural since the sampling stops once an initial trajectory which completes the path between Astrobee and the desired state is found. This trajectory is considered sub-optimal, but if the LQR-RRT* continued sampling the space through time, trajectories obtained would reach asymptotic optimality with respect to the cost function. For real-time hardware testing, computational efficiency is necessary for Astrobee to perform in real-world conditions. Using the LQR-RRT* trajectory, an LQR shortcutted trajectory is found which takes a shorter amount of time to reach the desired target by taking shortcuts. The shortcutting method reduces the effects of randomness that occurs from sampling using LQR-RRT*. This final motion plan is than tracked using robust tube MPC which provides disturbance rejection control for the robot. For Fig. \ref{fig:Exp1theta}, note that the initial conditions for orientation of LQR-RRT* and LQR shortcutting differ from the initial condition of the controlled response of the Astrobee. This was an error in the algorithm which was resolved in further experiments, but the results obtained from the experiment still hold. In this case, the Astrobee would slew in response to the initial tracking error which would stabilize through time. Further discussion and comparisons to standard MPC is discussed in Subsection \ref{robusttrackingsection}.

\subsection{Experiment Set 2: Information-Aware Planning and Model Estimation}
Characteristic of system identification procedures (sysID) is a usually offline-planned purely excitation trajectory. Given a start and goal point, a criterion based on the nature and information content of the measured data is optimized in order to improve estimate accuracy. In contrast to sysID, RATTLE proposes an online trade-off between information-aware exciting motion and goal-directed motion. The online re-planning capabilities allow inclusion of latest model parameters as well as dynamic obstacles in the environment, resulting in more reactive and optimal plans. At the same time, RATTLE presents the ability to turn off exploratory motion once estimate covariance drops below a certain value, potentially conserving fuel and energy. The question then arises, whether an adjustable amount of information awareness can provide sufficient excitation for adequately estimating the parameters. The results presented in this section demonstrate the capabilities of RATTLE juxtaposed with that of a general sys ID procedure, more specifically in what concerns ability to lend enough excitation for parameter convergence.

\begin{figure}[h!]
    \centering
	\includegraphics[width=1\linewidth]{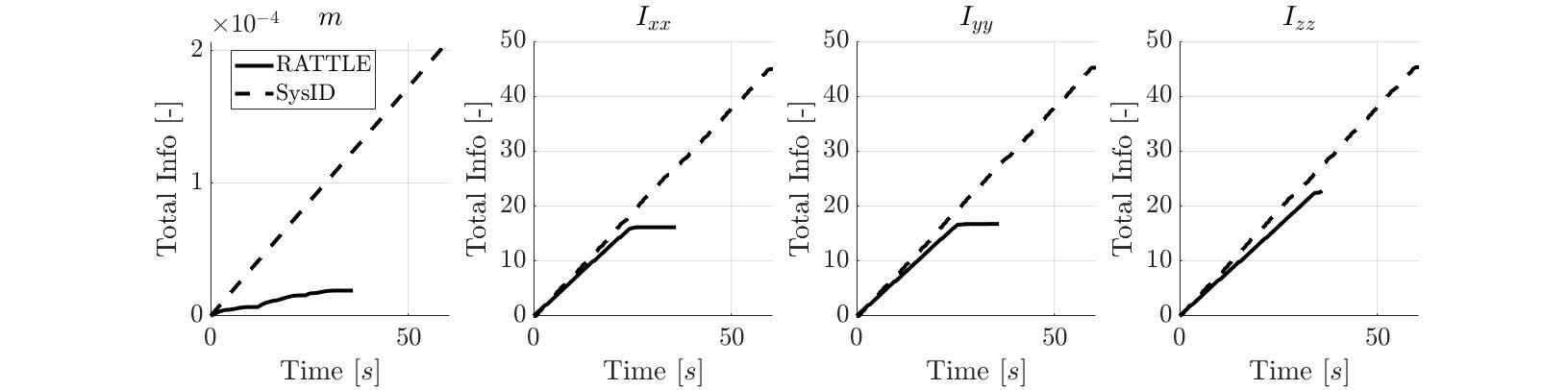}
   	\captionsetup{labelfont={bf}}
	\caption{Information content of the commanded purely informative and RATTLE local plans. This is the cumulative Fisher Information corresponding to the inertial parameter of interest.  While both trajectories start off with similar trends of `informativeness', particularly for the inertias, the information content of the RATTLE local plan shuts-off towards the end of the trajectory.   }
	\label{fig:info}
\end{figure}

Fig. \ref{fig:info} illustrates the information content in both cases, i.e., a purely excitation trajectory, and an information-aware trajectory using RATTLE. In the latter case, the emphasis on information weighting is relaxed as the estimate covariance decreases, causing the information content to plateau. Note that the robot performed a final yaw maneuver to reach the goal point, resulting in greater information content for $I_{zz}$ for the information-aware case. The excitation trajectory ran for 60 seconds and was executed during an earlier microgravity test campaign, whereas the RATTLE results shown here were a part of the complete on-orbit assembly pipeline, with 40 seconds of RATTLE-based parameter learning during the 3 minute demonstration.

  \begin{figure}[h!]
    \centering
	\includegraphics[width=1\linewidth]{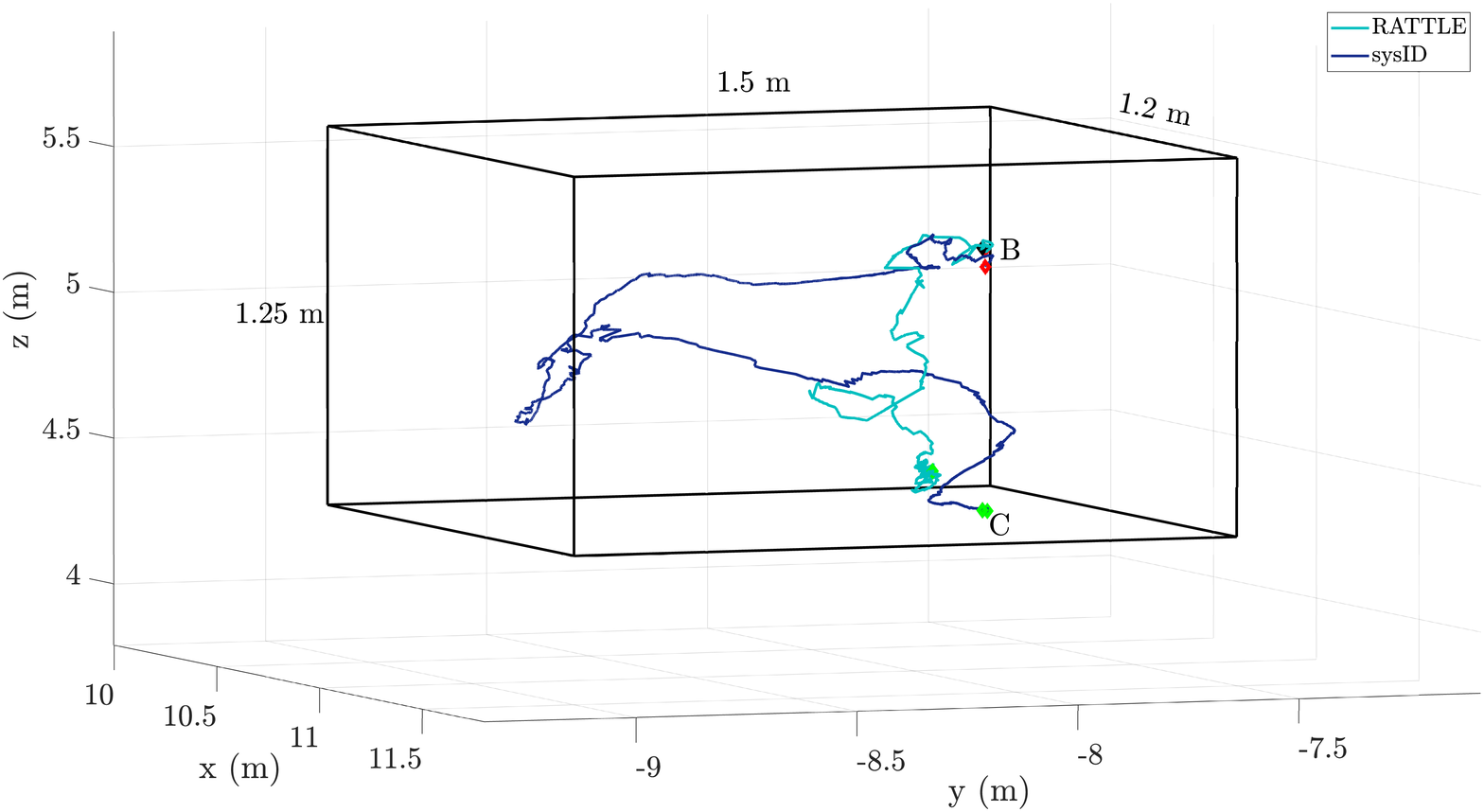}
   	\captionsetup{labelfont={bf}}
	\caption{Trajectories executed in microgravity, sysID and RATTLE. Black lines denote the space constraints for the planners. The red and green diamond markers denote the points B and C in 3D space. The RATTLE plan seems to have skipped regulation at point C, regulating at the last node of the RRT global plan instead.}
	\label{fig:3d_plan}
\end{figure}

   \begin{figure}[h!]
    \centering
	\includegraphics[width=1\linewidth]{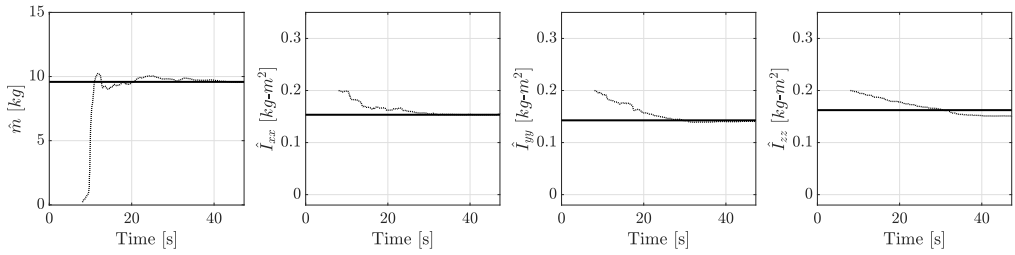}
   	\captionsetup{labelfont={bf}}
	\caption{Inertial parameter estimates for the RATTLE case. Since the effects detailed in Section \ref{sec:improving} were identified post microgravity-testing, the results shown here are obtained offline by applying the improved estimation algorithm on replayed microgravity test data. }
	\label{fig:estim}
\end{figure}

Tasks such as on-orbit assembly could benefit from the RATTLE algorithm's ability to replan and reduce parametric uncertainty online. As illustrated in Fig. \ref{fig:test13}, assembly tasks could be cyclic in nature, with the robot collecting and transporting payload to its assembly location and precisely depositing it. In this context, parametric uncertainty and the resulting inability to predict robot behaviour could potentially compromise safe and precise execution of the assembly task. While robust control approaches could offer robustness guarantees against known bounded uncertainties, these approaches are often cautious and do not take into account recent system/uncertainty information. On the other hand, estimating uncertainty by performing full system identification for each assembly component could be redundant and uneconomical in terms of fuel and time. To that end, RATTLE provides a reactive approach of online uncertainty reduction and incorporation of the latest system and environment information in subsequent optimal motion and robust control plans. Besides, as illustrated by the estimation plots in Fig. \ref{fig:estim}, sufficient information gain can be amassed on-the-fly without interrupting the primary task to perform full system ID. Additionally, the provision of online replanning is especially relevant in a single or multi-agent autonomous assembly scenario, where frequent online replanning might be needed as the assembled structure grows.

\begin{table}[h!]
  \captionsetup{labelfont={bf}}
  \centering
  \begin{tabular}{|c|c|c|}
    \hline
   Parameter & \% error & \% cov. reduction\\
    \hline
   mass & 0.39 & 99\\
    \hline
    $I_{xx}$ &  0.91 & 68\\
    \hline
    $I_{yy}$ & 1.35 &  49\\
    \hline
    $I_{zz}$ & 6.92 & 25\\
    \hline
  \end{tabular}
  \label{tab:est_breakdown}
    \caption{Percent error of with respect to the true values of the inertial parameters and percent covariance reduction for the estimates shown in Fig.\ref{fig:estim}}
\end{table}

\subsubsection{Overcoming Measurement Difficulties to Improve Estimation Accuracy}\label{sec:improving}

Latency, localization outliers, and input saturation were some key elements that directly impacted the accuracy of inertia estimation. The issues of measurement data latency and input saturation, which were more pronounced on hardware, were identified at the time of simulation testing. On the other hand, the effect of localization outliers and scaling was discovered after the microgravity test sessions. This section briefly explains the effects and the algorithmic components put in place to deal with them. 

\begin{itemize}
\item The parameter estimation algorithm received the robot states and commanded control inputs via the relevant ROS topics, denoted here as CTL for the control inputs, and EKF for the localization data. The expected publication frequency of both these topics is 62.5 Hz. However, as Fig. \ref{fig:freq} illustrates, latency occurred regularly on the resource-constrained hardware; the initial decrease in frequency is thought to arise from launching custom nodes and global plan computation. A time-stamp ordering algorithm was therefore implemented to synchronize data from the two sources before being passed on to the sequential estimator.
 \begin{figure*}	[h!]
 	\centering
 	\includegraphics[width=0.50\linewidth]{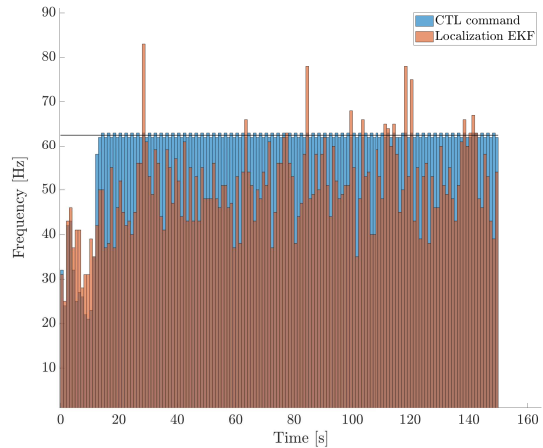}
 	\captionsetup{labelfont={bf}}
 	\caption{A histogram illustrating the frequency of the commanded control signals (CTL) and the localization estimates (EKF) on Astrobee hardware during a sample microgravity experiment. The bins correspond to the number of signals sent out per second (Hz). Note that the expected frequency of both signals is 62.5 Hz, as shown by the black line, but computation latency results in signals being dropped or delayed. }
 	\label{fig:freq}
 	
 \end{figure*}

 \item As discussed in Section \ref{sec:mixer}, the design of Astrobee's FAM is such that the actual actuation could be quite different from the commanded values. Tight actuation constraints to prevent frequent force and torque saturation coupled with a custom module dubbed \texttt{inverse\_fam} were put in place to deal with this issue. The \texttt{inverse\_fam} module uses the nozzle opening angles (published by default via a dedicated ROS topic), and in a non-trivial inverse mixer process obtains the actually applied post-saturation forces and torques at each time step.
 
\item Astrobee's known mass estimates are regarded as ``fairly accurate'', therefore further investigation of the measurement data was carried out when the obtained estimates did not match the expected values. To troubleshoot the estimation algorithm further, the projected linear accelerations, i.e., those found by dividing the commanded forces by the known mass of Astrobee, were compared to the measured accelerations, both quantities in body frame. A ``scaling effect'' was observed on the accelerations, with the measured accelerations being \textit{higher} than the projected ones, considering the known estimate of Astrobee's mass as valid. This effect, observed in data recorded across all experiments, manifested differently for each axis. While the reason for these ``stray accelerations'' is still being investigated, the parameter estimator treats this as a known linear model in addition to Astrobee's dynamics. Linear least squares was used for calibration of the scaling parameters using micro-gravity data. The scaling factor per axis is denoted as $s_i$, where $i = \{x,y,z\}$. Measured linear accelerations recorded over all RATTLE on-orbit experiments were stacked per axis, resulting in vectors $\mathbf{a}_i$, where $i = \{x,y,z\}$. Similarly, the force vectors over multiple experiments were collected as $\mathbf{f}_i$. The known true mass of Astrobee is denoted as $m$, $m = 9.58\ [\text{kg}]$. Let $\theta = 1/m$ and $\theta' = 1/s$. This formulation of the scaling parameter makes it straightforward to incorporate this effect post-calibration in the sequential estimation framework of Section \ref{sec:param_est}.
\begin{equation}
\theta'^*_i = argmin\vert\vert \mathbf{f}_i(\theta + \theta'_i) - \mathbf{a}_i\vert\vert_2^2
\end{equation}
\begin{equation}
\theta'^*_i = (\mathbf{f}_i^T\mathbf{f}_i)^{-1}\mathbf{f}_i^Ta_i - \theta
\end{equation}

 \item Owing to its minimization of the $L_2$ norm of the residuals, the least squares estimation is particularly sensitive to outliers, which can skew the fit. In the case of the ReSWARM micro-gravity experiments, sudden jumps in Astrobee's localization were a key reason for the loss of accuracy in the estimation procedure. Fig. \ref{fig:outlier} shows one such localization outlier as seen in the angular acceleration. Note that the angular accelerations were computed by numerical differentiation of the angular velocity data, resulting in amplification of the measurement noise.
In the recursive least squares (or Kalman filter) formulation, Section \ref{sec:param_est}, the innovation, or the difference between the observed and the predicted measurements based on the current estimate are given as

\begin{figure*}[h!]
    \centering
    \includegraphics[width=0.45\linewidth]{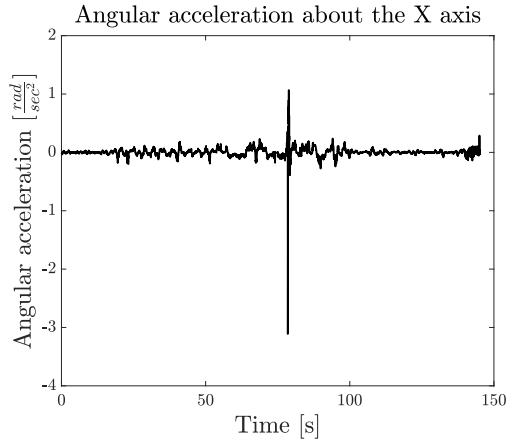}
    \captionsetup{labelfont={bf}}
    \caption{Angular acceleration showing localization outliers during a microgravity run. Apart from the jumps seen in the plot at 78 seconds, some other localization drifts can be seen. Fast motions that cause a loss of visual features, as well as external disturbance forces are believed to cause these effects. } 
    \label{fig:outlier}
\end{figure*} 

\begin{equation}
    \gamma_{k} =  \tilde{\mathbf{y}}_k - \mathbf{H}_{k}\hat{\theta}_{k-1}
\end{equation}

and the innovation covariance, $E\left[\gamma_{k},\gamma_{k}^T\right]$, is

\begin{equation}
    \mathbf{S}_{k} = \mathbf{H}_{k}\mathbf{P}_{k-1}\mathbf{H}^T_{k} + \mathbf{W}
\end{equation}

If the assumption of randomly distributed Gaussian measurement noise from eq. (\ref{eq:estim_formulation}) holds, then the normalized sum of innovations $\gamma_{k}S^{-1}_{k}\gamma_{k}^T$,  corresponds to the Mahalanobis distance between the expected measurement based on the previous estimate and the actual measurement \cite{mirzaei2008kalman}. Based on the critical value of the $\chi^2_l$ distribution at the desired confidence level, a probabilistic threshold can thus be set for $\gamma_{k}S^{-1}_{k}\gamma_{k}^T$ to determine measurement reliability. Here, $l$ is the number of degrees of freedom, corresponding to the number of independently sampled normal distributions. In this case, $l$ is equal to the number of measurements, $l = 6$ (linear and angular accelerations), as they are considered independent and identically distributed. The value corresponding to the probability level of 0.99 is chosen as the threshold $\nu$, i.e., $P(\gamma_{k}^TS^{-1}_{k}\gamma_{k}\leq \nu) = 0.99 $. 
\begin{equation}
\gamma_{k}^TS^{-1}_{k}\gamma_{k} \in \chi^2_l
\end{equation}
For outlier rejection, if the result of the test was higher than a threshold $\gamma_{k}^TS^{-1}_{k}\gamma_{k} > \nu$, then the measurement was rejected and the update steps from eq. \ref{eq:est_update} were not performed. 
\end{itemize}

\subsection{Experiment Set 3: Robust Tracking}\label{robusttrackingsection}
\begin{figure}[hbtp!]
  \centering
  \includegraphics[width=0.8\linewidth]{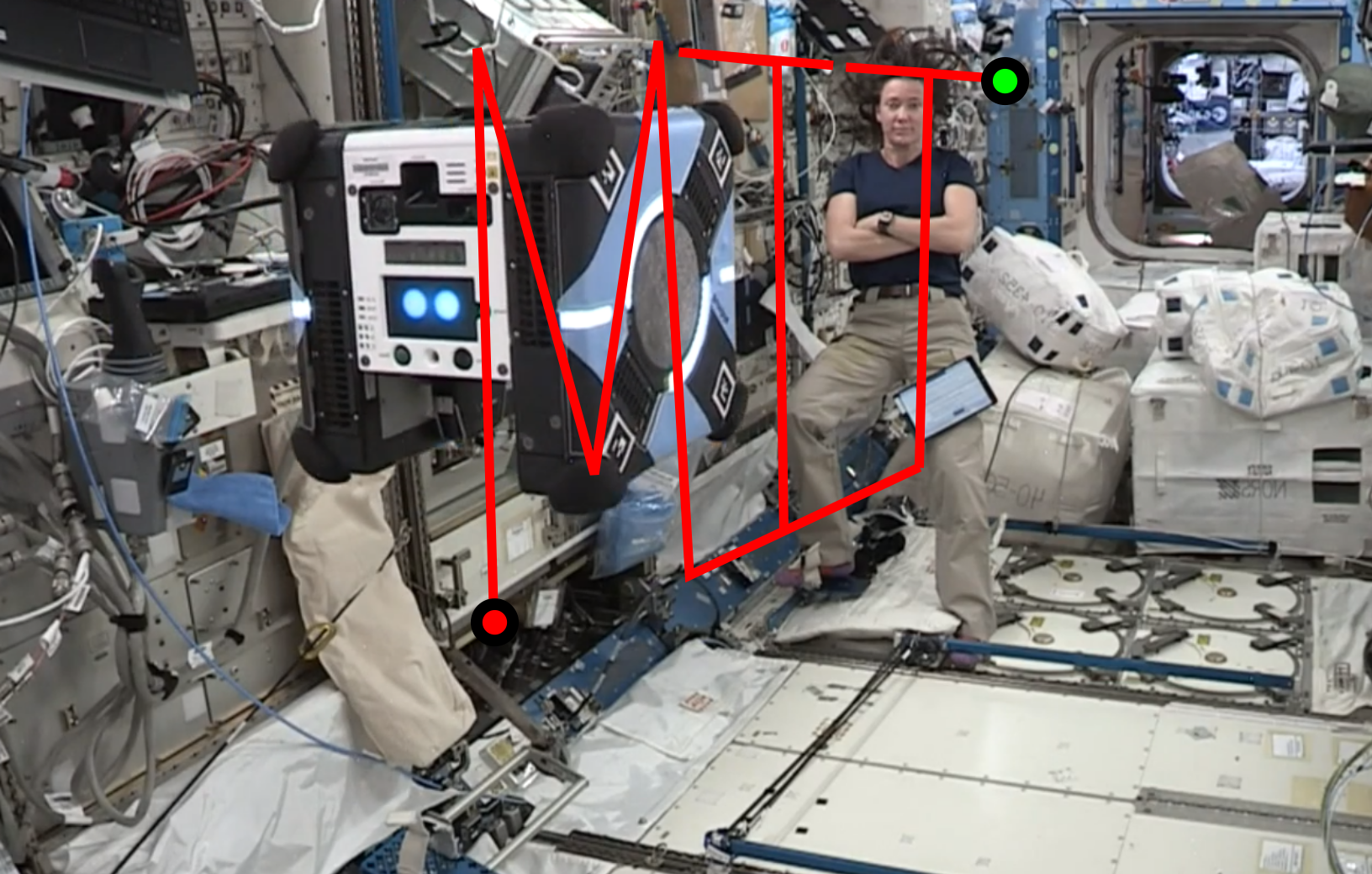}
  \caption[The MIT reference trajectory.]{The tracking task's reference trajectory for robust tube MPC comparison.}
  \label{fig:mit}
\end{figure}

In addition to the on-orbit assembly scenario, a separate, dedicated set of experiments sought to demonstrate the performance benefits of robust tube MPC for trajectory tracking. Here, parametric uncertainty was ignored (the correct model was supplied). Astrobee was tasked with tracking a translational reference trajectory, with the only error sources present those inherent in Astrobee's operation.\footnote{While somewhat speculative, Astrobee's chief disturbance contributor is its localization system, perhaps followed by the accuracy of its mixer. There are a variety of other error contributors, such as: an always-on cooling fan; air vents on the ISS; a rampdown/rampup period for Astrobee's impeller system; and others.} The tracking task is shown in Fig. \ref{fig:mit} (note that the image is reversed along the ISS y-axis).

\begin{figure}[hbtp!]
  \centering
  \includegraphics[width=1.0\linewidth]{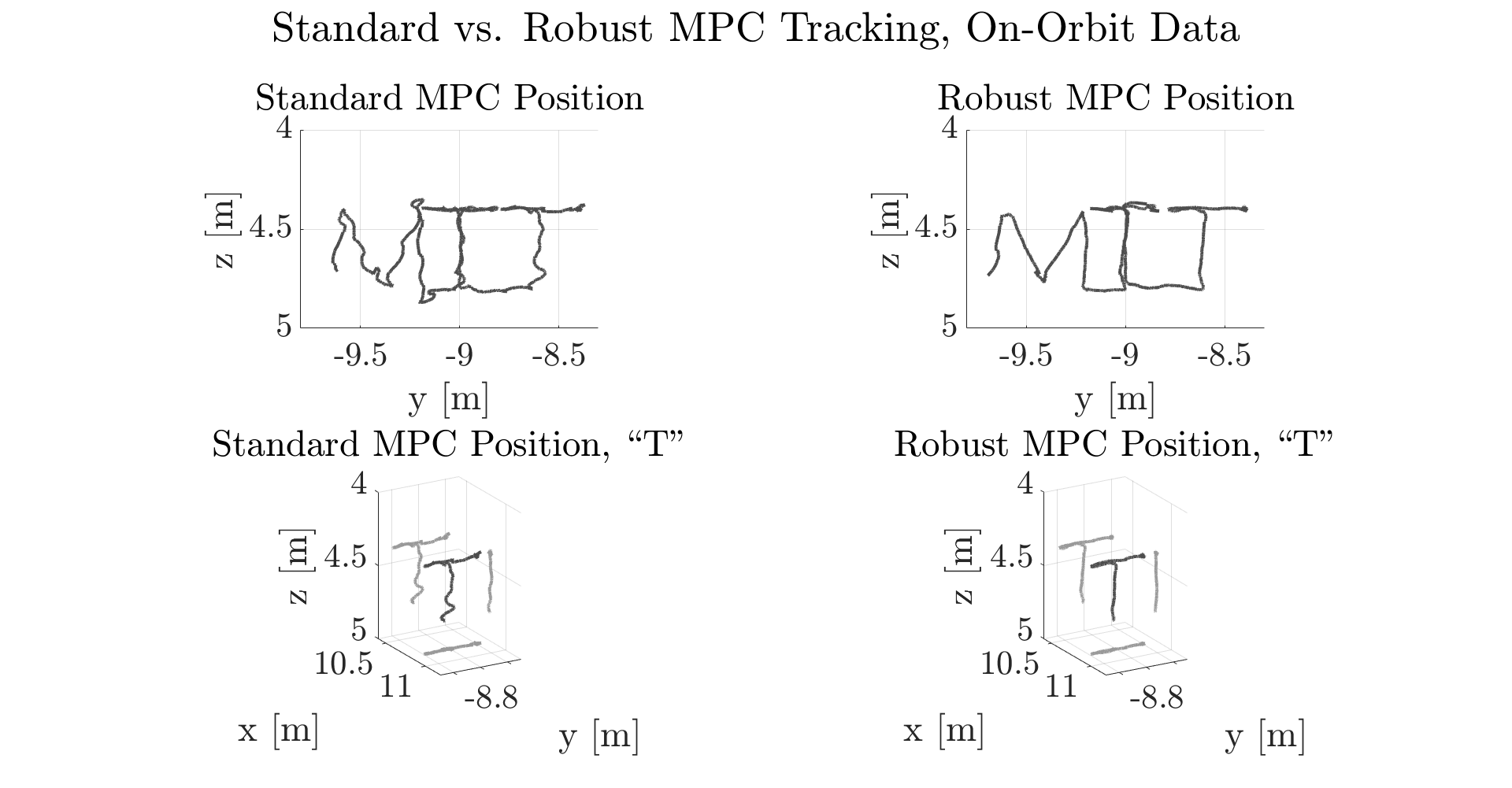}
  \caption[Tracking the MIT reference trajectory.]{Tracking of the reference trajectory with standard MPC (left) and robust tube MPC (right), with the same cost function tuning values. A 3D view of the ``T'' portion of the trajectory is shown below. Note that a portion of the initial trajectory has been clipped during a transient period where Astrobee's impeller ramps down when swapping controllers.}
  \label{fig:mit_robust_comp}
\end{figure}

\begin{figure}[hbtp!]
  \centering
  \includegraphics[width=1.0\linewidth]{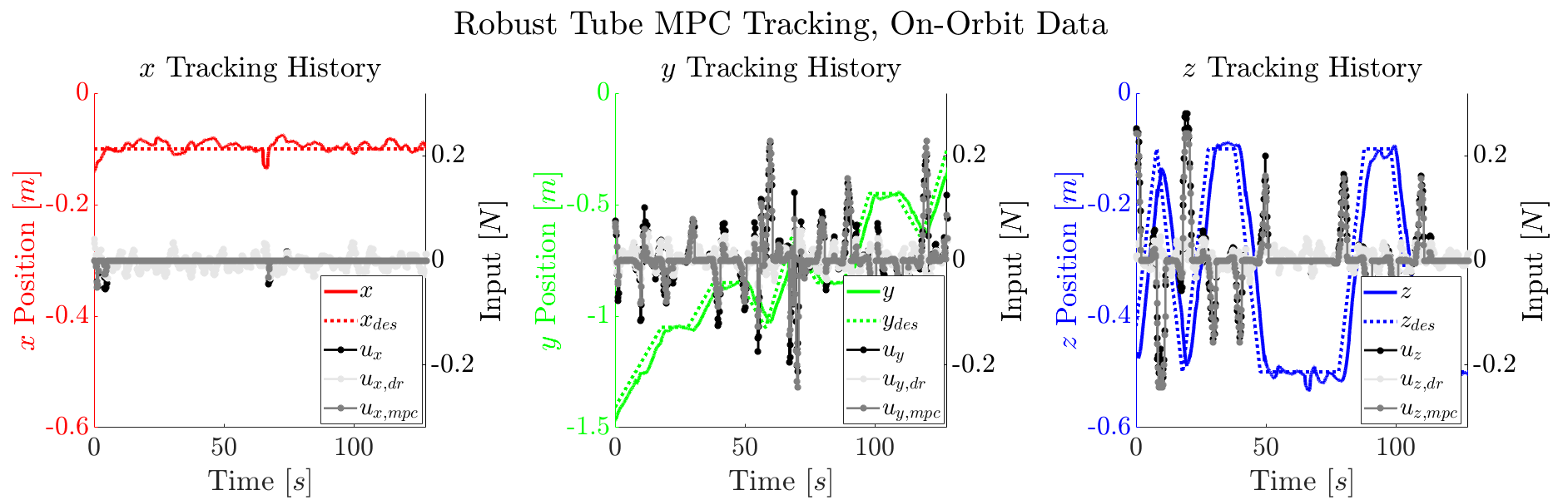}
  \caption[On-orbit robust tube MPC tracking.]{Translational tracking performance for the x-, y-, and z-axes, with robust tube MPC input components also shown. Robust tube MPC's total input is the combination of a tightly-constrained nominal MPC and a disturbance rejection ancillary controller.}
  \label{fig:robust_tracking}
\end{figure}

The results of the tracking task are shown in Fig. \ref{fig:mit_robust_comp}. Robust tube MPC is more ``sluggish'' due to its tightened nominal constraints, but is generally better in counteracting disturbances due to its disturbance rejection controller. This is particularly obvious in Fig. \ref{fig:robust_tracking}, where inputs are plotted alongside position tracking performance. Here, the nominal (grey), disturbance rejection (white), and total (black) control inputs have been separately recorded from on-orbit data. Generally, the nominal MPC portion is relatively inactive if there are no major changes in trajectory tracking, particularly evident for the x-axis tracking which is unchanged for the duration of the trajectory. As seen in the z-axis tracking, nominal MPC activates to provide horizon guidance once upcoming trajectory changes come within view. Throughout, disturbance rejection control activates to counter disturbance sources. This experiment is a good example of robust tube MPC's ability to provide a form of robustness in its tracking task. As shown later in Section \ref{sec:speedy}, it is also feasible to run on resource-constrained hardware.

\subsubsection{Robust Tracking Results for the On-Orbit Assembly Scenario}
As mentioned in Subsection \ref{sec:approach}, the on-orbit assembly scenario involves 3 segments which the Astrobee moves from waypoints (a), (b), and (c) following the algorithmic elements found in Table \ref{tab:my_label}. Additionally, a static obstacle is set between waypoints (a) to (b) and (c) to (a) to model the physical space where ``construction components'' reside. 

\begin{figure}[hbtp!]
  \centering
  \includegraphics[width=1.0\linewidth]{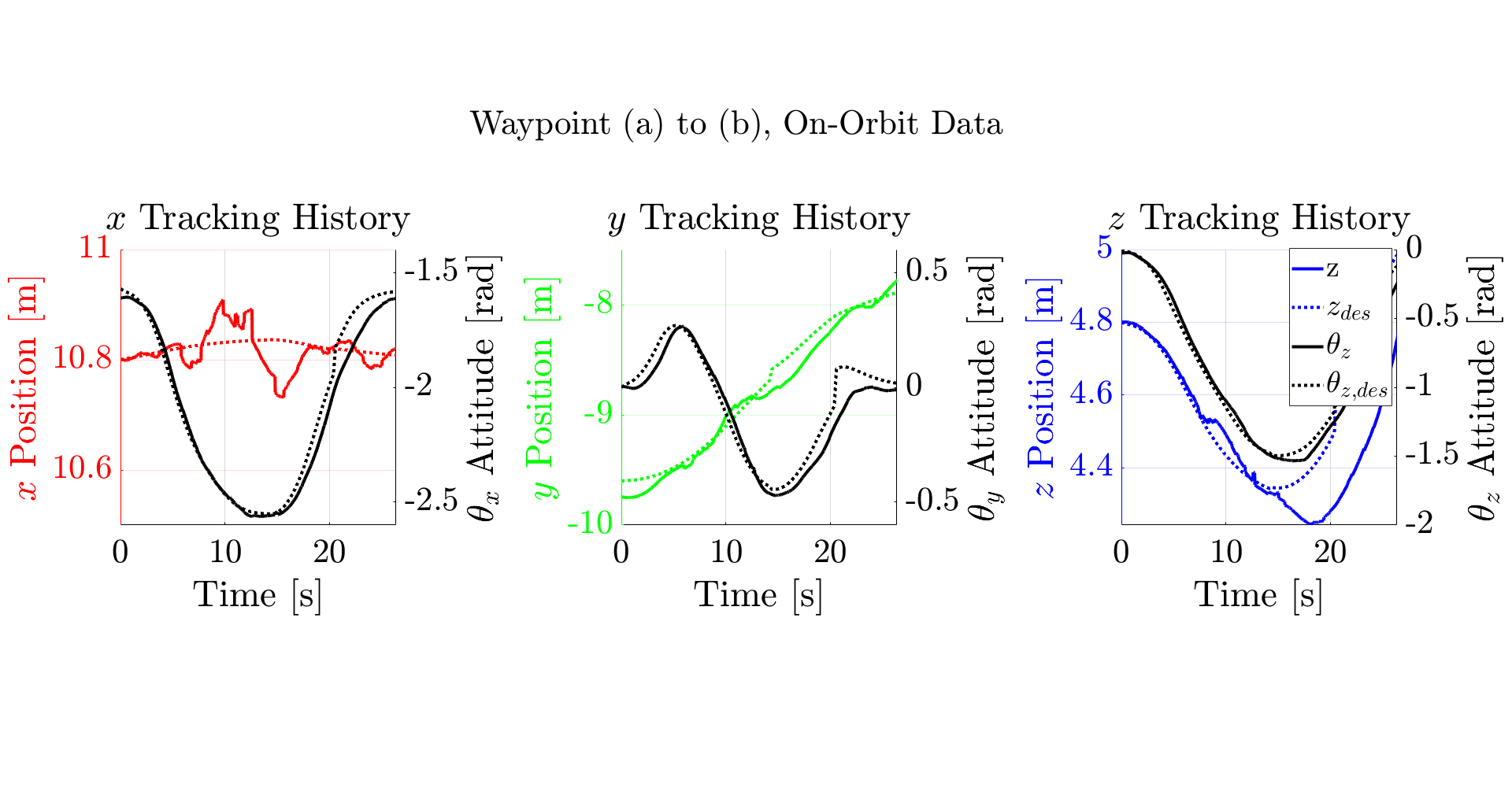}
  \caption[]{Translational and attitude histories of the Astrobee tracking a LQR-RRT* plan en route from (a) to (b).}
  \label{fig:atobxyz}
\end{figure}

For the first segment, the Astrobee moves from the construction site to the storage area to ``manipulate" a component for construction. At this part of the scenario, the system model knowledge is adequate with no inertial uncertainty, so standard MPC is used. The Astrobee obtains collision-free trajectories (using LQR-RRT* and LQR shortcutting) and tracks the motion plan which is shown in Fig. \ref{fig:atobxyz}.

\begin{figure}[hbtp!]
  \centering
  \includegraphics[width=1.0\linewidth]{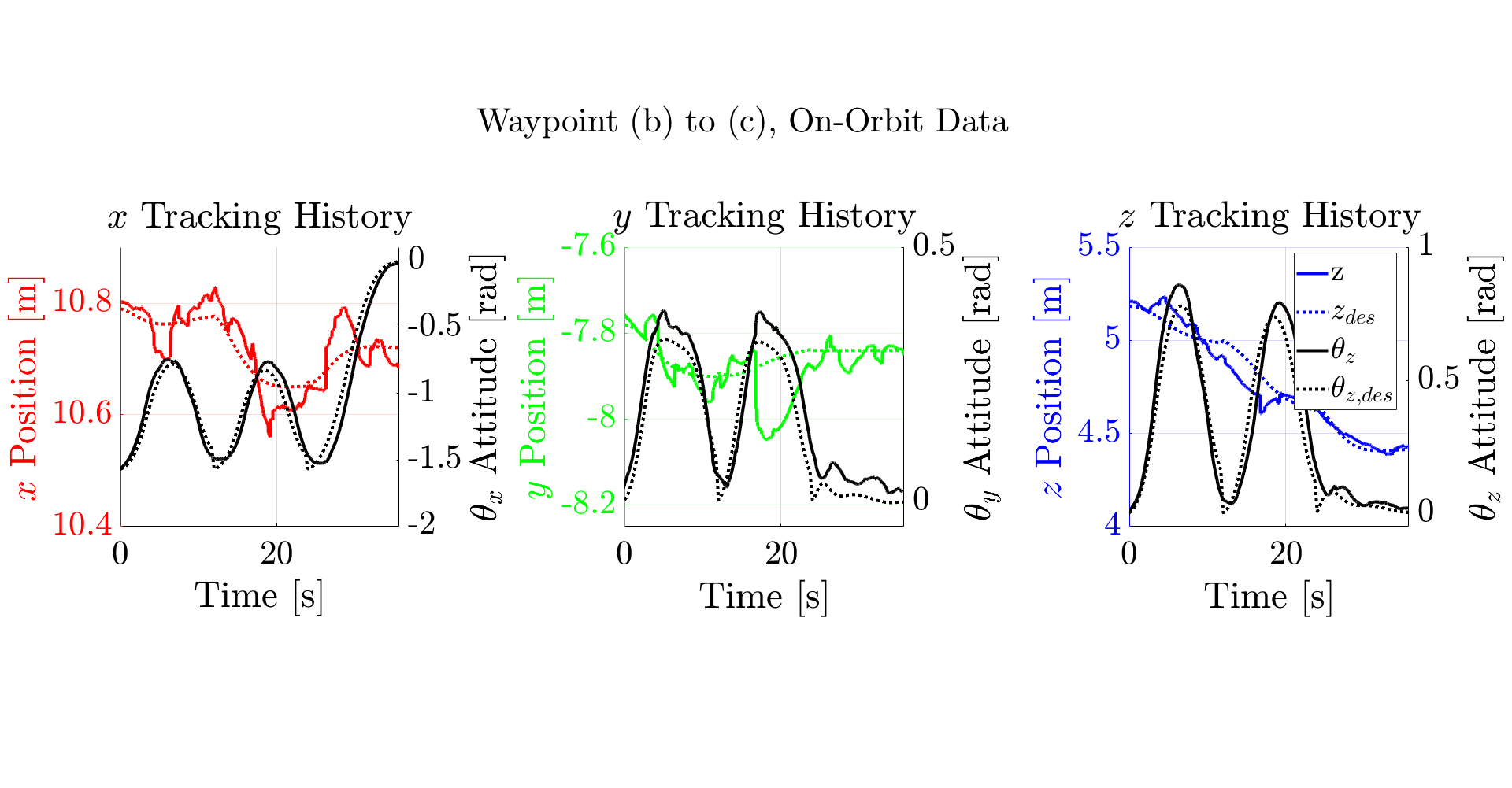}
  \caption[]{Translational and attitude histories of the Astrobee tracking an information rich plan en route from (b) to (c).}
  \label{fig:btocxyz}
\end{figure}

The second segment involves the Astrobee moving from storage area (b) to a safe area (c). During this segment, it is assumed that the Astrobee has already manipulated the object, thus the system model knowledge of the inertial parameters are poor. The RATTLE algorithm is used to plan and estimate the inertial parameters while updating the system model to better the controlled response. Fig. \ref{fig:btocxyz} shows the information rich trajectory and the robust tube MPC controlled response using the RATTLE algorithm. Fig. \ref{fig:3d_plan} shows the real-time planning of the RATTLE motion planner which is used in conjunction with the robust tube MPC controller and sequential least squares to update the system model. Once Astrobee reaches waypoint (c), Astrobee has a sufficient model of the learned inertial parameters to move onto the next segment.

\begin{figure}[hbtp!]
  \centering
  \includegraphics[width=1.0\linewidth]{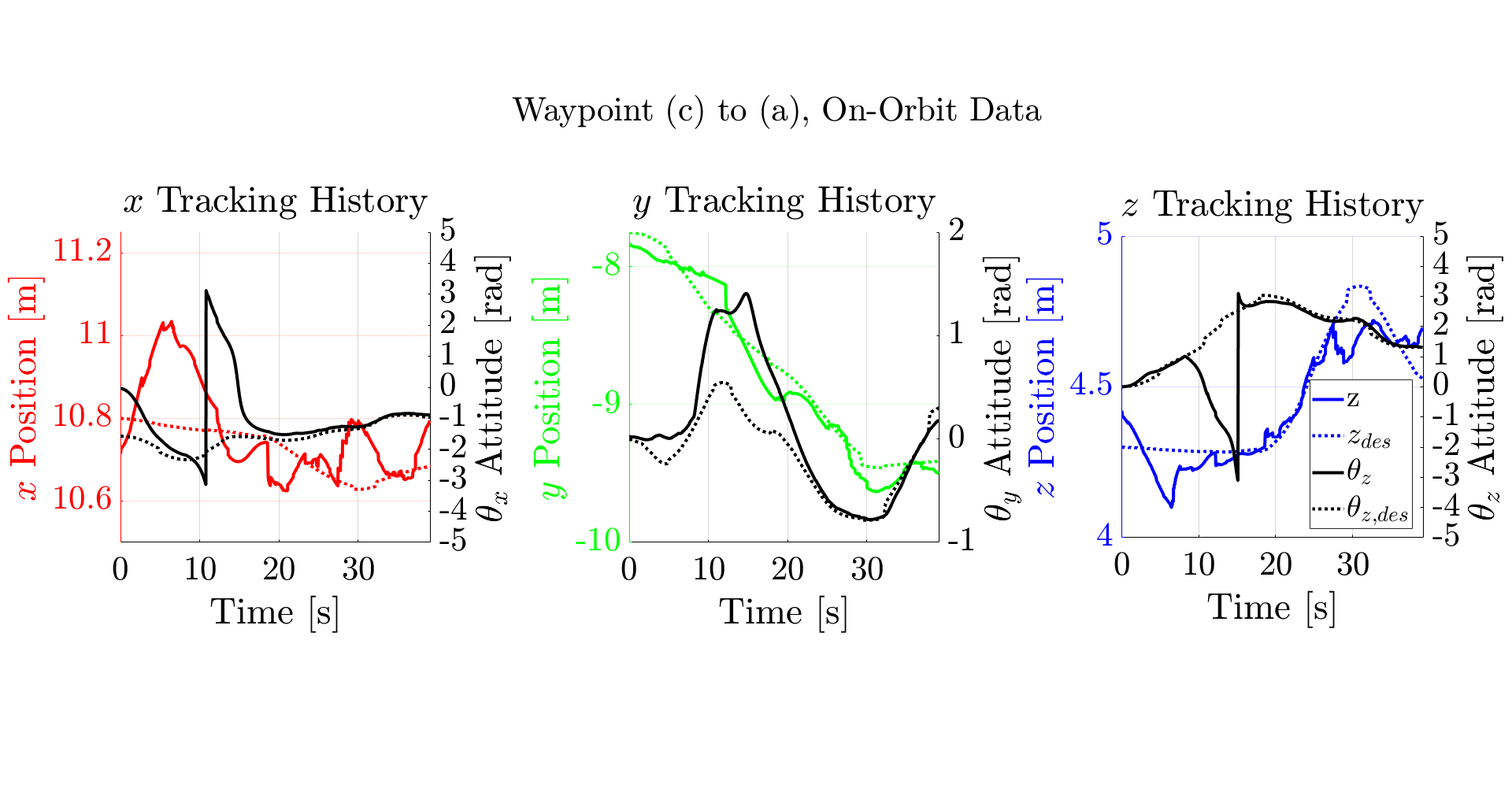}
  \caption[]{Translational and attitude histories of the Astrobee tracking a LQR-RRT* plan en route from (c) to (a). Note jumps due to the localization system, discussed further in this section.}
  \label{fig:ctoaxyz}
\end{figure}

With the inertial parameters learned, the Astrobee system model is sufficient, and the Astrobee moves from the safe area (c) back to the construction site (a). A high-level collision-free motion plan is formed using LQR-RRT* and LQR shortcutting to avoid the obstacle between the two waypoints. With the learned system model, uncertainty is still present, thus robust tube MPC is used to mitigate this disturbance as it tracks the motion plan. Fig. \ref{fig:ctoaxyz} shows the motion plan and the robust tube MPC response for this segment. 

The localization error played a significant role in the tracking accuracy of the Astrobee system. This is much more apparent in the (c) to (a) segment. At position (a), the Astrobee is much closer to features allowing for more accurate localization estimates, but the position at (c) is further from these features which increases the variability in localization error. Comparing Fig. \ref{fig:ctoaxyz} to \ref{fig:atobxyz}, the tracking error in all three directions is much more significant in the (c) to (a) segment. The ``jumps" in the tracking error from Astrobee's rotational motion is due to this localization error. The increase in localization error is due to the decrease in feature count which is attributed to distance from the features, physical changes to the experimental module that can hide features, and rotational velocity of the Astrobee when collecting features. These contributions to the localization error were mitigated as much as possible during the experiment, yet the localization error produced a significant effect on the tracking response for the experiment. 

\subsection{Planning and Control Speeds on Resource-Constrained Hardware}
\label{sec:speedy}

\begin{figure}[hbtp!]
  \centering
  \includegraphics[width=1.0\linewidth]{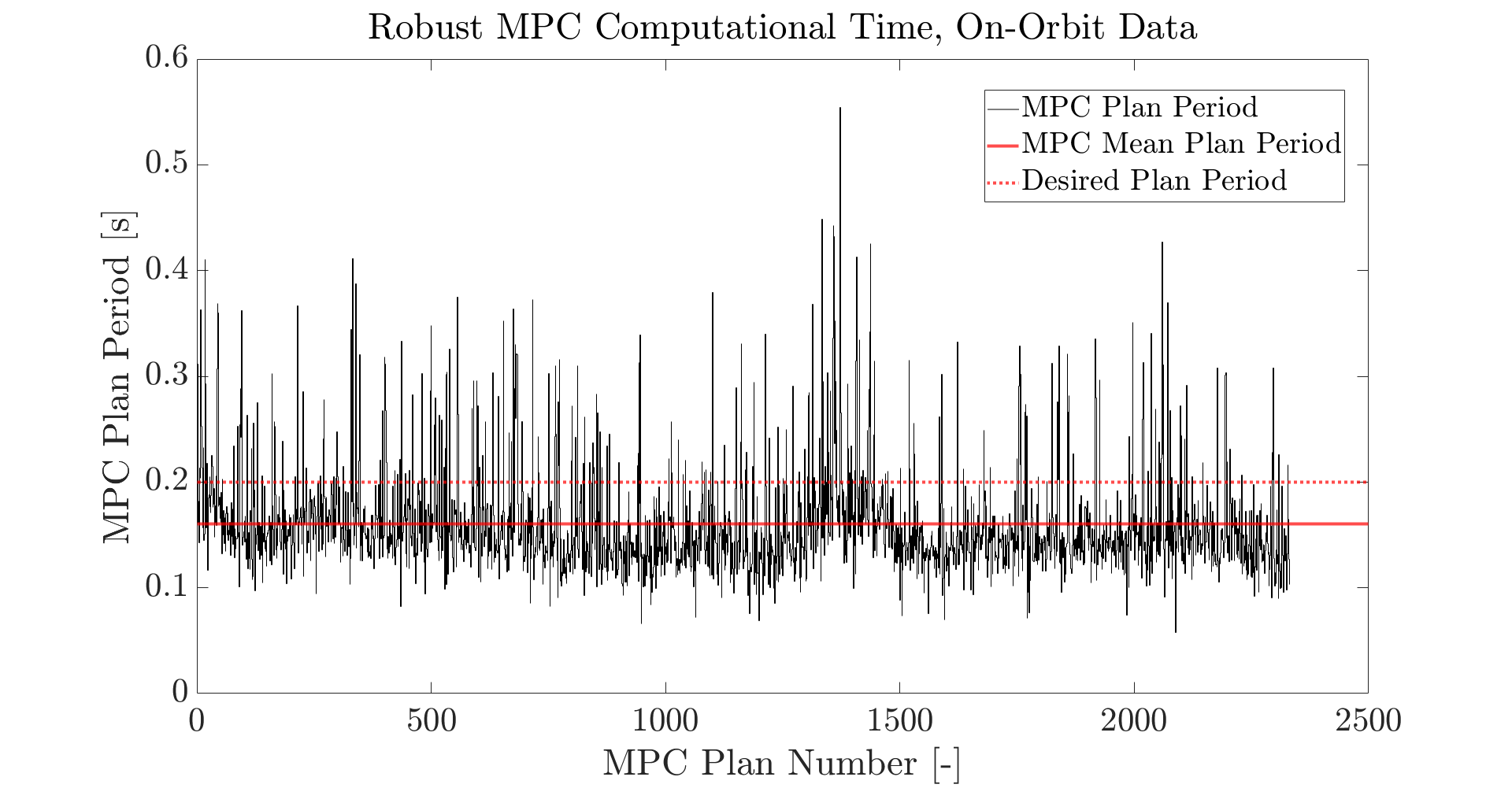}
  \caption{Typical planning periods for robust tube MPC onboard the Astrobee free-flyer, shown here for a multi-waypoint maneuvering task.}
  \label{fig:comp_time}
\end{figure}

Finally, online planning and control computational speeds are considered aboard Astrobee, an important consideration when discussing real-time use on hardware. Planning/control periods and approximate hardware-demonstrated values are shown in Fig. \ref{tab:comp_speed}. For example, the computational speed of an example run using the robust tube MPC is shown in Fig. \ref{fig:comp_time}. While outlier periods do exceed the desired computational period, average computational time is within the desired $0.2\ $[s] computational period ($0.1604 \pm 0.0503\ $[s]). Exact processor details are provided in \cite{Smith2016}; significant effort went into ensuring components met their expected rates. Planning and control periods had to be carefully chosen in a tradeoff between increasing optimality for longer horizons, the frequency of incorporating desired updates for replanning (especially for the local planner), and the computational realities of the Astrobee processors. A notable difference on-orbit compared to ground testing was the increased load for the default localization system, which has to contend with a larger map and a greater number of features, and other flight software components, resulting in an approximately $\times 2$ slowdown compared to ground tests.

\begin{table}[]
  \caption[Replanning periods.]{Replan periods and approximate actual computational time taken for computation on Astrobee's Snapdragon processors on-orbit.}
  \centering
  \begin{tabular}{|c|c|c|}
    \hline
    Component & Replan Period [s] & Hardware Speed [s]\\
    \hline
    \hline
    Global Planner (LQR-RRT*) & as-needed & 110 - 125\\
    \hline
    Global Planner (kino-RRT) & as-needed & $0.5 - 2$\\
    \hline
    Local Planner (RATTLE only) &  12 & $2 - 7$\\
    \hline
    Controller & 0.2 & $0.05 - 0.2$\\
    \hline
  \end{tabular}
  \label{tab:comp_speed}
\end{table} 

\paragraph{Simulation-Based Benchmark Data for Optimization-Based Solvers}
RATTLE's local planner and robust tube MPC controller ultimately rely on optimization solvers to handle online computation of reference trajectories and control inputs, respectively (see Section \ref{sec:methods}). Simulation benchmark details of these components are provided for a better understanding of their performance and their comparison to similar optimization-based solutions.

The information-aware local planner uses ACADO \cite{houskaACADOToolkitAnOpensource2011}, a software toolkit for solving optimal control problems. ACADO relies on a direct multiple shooting discretization with an SQP solution method, with QPOASES as the default QP solver  \cite{ferreauQpOASESParametricActiveset2014}. 
Well-suited for optimal control problems, ACADO uses a real-time itreation (RTI) scheme, whereby the quadratic approximation to the true NLP is constructed in separate preparation and computation phases for each SQP iteration, allowing problem preparation to occur before measurements are received. Multiple iterations may be performed if greater local convergence accuracy is desired---this is the approach taken by RATTLE, performing as many iterations as possible until convergence criteria fall below a designated threshold or computation time runs short. The KKT conditions are used as an approximation of convergence quality. Meanwhile, the robust tube model predictive controller also uses QPOASES, with CasADi as the front-end for specifying the optimal control problem \cite{anderssonCasADiSoftwareFramework2019}. The robust tube MPC benchmark uses on-orbit data from a sample reference trajectory, with simulated disturbance error sampled from a uniform distribution. Ultimately, both solution methods rely on the online solution of quadratic programs---in the case of the local planner, as the backend to a nonlinear programming solution method.

Table \ref{tab:solvers} compares some benchmarking data for each of these solvers for simulation data of plans with horizons equivalent to that used in later hardware demonstrations ($N=40$ for local planning, $N=5$ for robust control). KKT tolerance is used as a convergence benchmark for SQP. Five SQP iterations are performed for the local planner, near the maximum permissible iterations on hardware while reliably remaining within a replanning period of 12 [s]. The local planner benchmark includes information weighting on moments of inertia and mass for an approximately 0.5 [m] offset waypoint planning task using the 6DOF Newton-Euler dynamics. The tube MPC, meanwhile, is solved exactly at each call using QPOASES for a sinusoidal tracking task for the double integrator dynamics. Benchmark data uses the same typical ``laptop'' hardware used for global planning shown above. Note that planning/control periods and rates are explicitly crafted to be sufficient for use on Astrobee's more limited processors, detailed in Section \ref{sec:speedy}.

\begin{table}[]
  \caption[Local planner and controller solve time information.]{Computation timing on representative benchmark problems for the local planner and controller in simulation.}
  \centering
  \resizebox{\textwidth}{!}{
    \begin{tabular}{|c|c|c|c|c|c|}
      \hline
      Module & \thead{Problem Size\\{[}dec. vars.]} & Problem Class & Solver(s) & \thead{Benchmark\\Solve Time [s]} & \thead{Convergence Data} \\
      \hline
      \hline
      Local Planner (N=40) & 240 & NLP & ACADO RTI SQP solver\tablefootnote{Note that ACADO uses a direct multiple shooting approach, where the full state need only be provided at a number of ``shooting nodes.'' This 40-timestep solution uses only 6 input variables, hence 240 QP decision variables.}, QPOASES& 0.929 & $4.53\times10^{-4}$ (KKT stationarity norm)\\
      \hline
      Controller (N=5) & 51 & QP & QPOASES (and qrqp)& 0.00849 & 0.0 (exact convergence)\\
      \hline
    \end{tabular}}
    \label{tab:solvers}
  \end{table}

\section{Conclusion}
\label{sec:conclusion}
Robotic spacecraft executing on-orbit operations must address autonomy challenges such as collision-avoiding motion planning in dynamic environments and ensuring control accuracy in the presence of changing system models. To that end, this work develops planning, control, and model learning frameworks for tackling these challenges, and presents novel microgravity experimentation results from the ReSWARM flight experiments on NASA's Astrobee free-flyer on the International Space Station. Considering an on-orbit assembly scenario, RElative Satellite sWarming and Robotic Maneuvering (ReSWARM), demonstrated the following novel technologies on robotic hardware in the ISS microgravity environment for the first time: (1) Sampling-based global planning taking into account system dynamics, (2) information-aware planning and model learning for on-orbit reconfiguration, and (3) tube-based robust tracking control in the presence of model uncertainty. Further, extensive hardware implementation details and challenges that future autonomous free-flyers will need to considered are presented.  

The demonstrated framework for microgravity testing provides the capabilities for safe, modular, close proximity operation. These strategies can be repurposed to other on-orbit autonomy tasks that require efficiency from a safety standpoint. By demonstrating these technologies on practical hardware tests, the framework can feasibly be incorporated in new emerging space robotic systems for larger and more complex on-orbit close proximity applications. Future work will entail physical manipulation of objects to further validate the on-orbit assembly demonstration, consideration of physical objects for real-time mapping and collision avoidance, and bringing the information-theoretic framework to a greater set of uncertain robots.

\subsubsection*{Acknowledgments}
This research was supported by an appointment to the Intelligence Community Postdoctoral Research Fellowship Program at Massachusetts Institute of Technology, administered by Oak Ridge Institute for Science
16 and Education through an interagency agreement between the U.S. Department of Energy and the Office of the Director of National Intelligence. Funding for this work was also provided by the NASA Space Technology Mission Directorate through a NASA Space Technology Research Fellowship under grant 80NSSC17K0077. This work was also supported by the Portuguese Science Foundation (FCT) grant PD/BD/150632/2020, the
LARSyS - FCT Plurianual funding 2020-2023, and an MIT Seed Project under the MIT Portugal Program. International Space Station (ISS) experiments were conducted under ISS National Lab user agreement UA-2019-969 as part of the Relative Satellite sWArming and Robotic Maneuvering (ReSWARM) investigations. The researchers would especially like to thank Brian Coltin, Marina Moreira, Ruben Garcia Ruiz, Ryan Soussan, Jose Benavides, Jose Cortez, Aric Katterhagen, and the rest of the Astrobee operations team at NASA Ames. Ames' stellar operations staff accommodated ground testing (including many remote test sessions), assisted with software questions, and generally collaborated to improve the Astrobee software experience. The authors gratefully acknowledge the support that enabled this research.

\bibliographystyle{apalike}
\bibliography{jfrExampleRefs, keenan_bib}  

\end{document}